%% file: main.tex
\documentclass[lettersize,journal]{IEEEtran}
\usepackage{amsmath,amsfonts}
\usepackage{algorithmic}
\usepackage{algorithm}
\usepackage{array}
\usepackage{textcomp}
\usepackage{stfloats}
\usepackage{url}
\usepackage{verbatim}
\usepackage{graphicx}
\usepackage{subcaption}
\usepackage{cite}
\usepackage{amssymb}
\usepackage{pifont}
\usepackage{dsfont}
\usepackage[table]{xcolor}
\usepackage{tabularx}
\usepackage{multirow}
\usepackage{threeparttable}
\usepackage{caption}
\usepackage{booktabs}
\usepackage{adjustbox}
\usepackage{xspace}
\usepackage{xcolor}
\usepackage{tikz}
\usepackage{multicol}
\usepackage{enumitem}
\usepackage[utf8]{inputenc}
\usepackage[most]{tcolorbox}

\definecolor{mygreen}{RGB}{0,200,0} 
\newcommand{\mygreen}[1]{\textcolor{mygreen}{#1}}
\definecolor{myred}{RGB}{200,0,0} 
\newcommand{\myred}[1]{\textcolor{myred}{#1}}
\newcommand{\ie}[1]{\textit{i.e.}}

\newtcolorbox{ClassBox}[1]{
    colback=gray!10,          
    colframe=black,           
    colbacktitle=black,       
    coltitle=white,           
    fonttitle=\bfseries\large, 
    arc=5pt,                  
    title=#1,                 
    boxrule=1pt,              
    left=10pt, right=10pt,    
    top=10pt, bottom=10pt     
}

\begin{document}

\title{Hierarchical Fine-Grained Aerial Object Detection}

\author{Yan Zhang$^\dagger$, Fang Xu$^\dagger$, Wen Yang, Gui-Song Xia$^*$
\thanks{$^\dagger$These authors contribute equally.}
\thanks{$^*$Corresponding author.}
\thanks{Yan Zhang is with the School of Computer Science, Wuhan University, Wuhan, 430072, China (email: zhangyan@whu.edu.cn)}
\thanks{Fang Xu and Gui-Song Xia are with the School of Artificial Intelligence, Wuhan University, Wuhan, 430072, China (email: xufang@whu.edu.cn, guisong.xia@whu.edu.cn)}
\thanks{Wen Yang is with the School of Electronic Information, Wuhan University, Wuhan, 430072, China (email: yangwen@whu.edu.cn)}
}

\markboth{Journal of \LaTeX\ Class Files}%
{Shell \MakeLowercase{\textit{et al.}}: A Sample Article Using IEEEtran.cls for IEEE Journals}


\maketitle

\begin{abstract}
Fine-grained aerial object detection, driven by the intrinsic granularity of real-world object categories, is crucial for advanced scene understanding in remote sensing.
Existing methods largely inherit the paradigm of coarse-grained object detection, relying solely on single-label supervision and thus struggling to distinguish model-level categories with subtle structural differences.
However, for each specific model (e.g., Boeing 787), structured prior knowledge such as attributes (e.g., wing shape) and hierarchies (e.g., plane$\xrightarrow{}$wide-body airliner) offers discriminative semantics across multiple granularities. 
Motivated by this, we present ExpertDet, a scheme that incorporates expert-informed cues to enhance fine-grained aerial object detection. 
Specifically, we design \textit{Vision-aware Masked Attribute Modeling} (VMAM), which aligns attribute semantics with visual structures by reconstructing randomly masked attributes from visual cues, enabling the detector to capture subtle structural distinctions. 
We further propose \textit{Hierarchical Visual Instance Promotion} (HierVIP), which builds a visual prototype tree based on hierarchical relations and imposes taxonomy-aware constraints to preserve cross-level semantic continuity while enhancing category discrimination.
Moreover, we curate a new fine-grained object detection benchmark for \underline{P}recise recognition of model-specific \underline{S}hips and \underline{P}lanes from aerial imagery, \textit{PSP}, covering 106 ship classes and 30 airplane models, respectively, featuring the most extensive collection of model-specific categories among existing aerial object detection datasets to date.
We benchmark state-of-the-art object detection algorithms on the \textit{PSP} benchmark. Extensive evaluation demonstrates that ExpertDet consistently outperforms other fine-grained competitors across hierarchy levels. The dataset, benchmark, and code are available at \url{https://nnnnerd.github.io/PSP-Benchmark/}.
\end{abstract}



\input{secs/intro}

\input{secs/related}

\input{secs/method}

\input{secs/dataset}

\input{secs/exp}

\section{Conclusion}
In this work, we present ExpertDet, a scheme that incorporates expert-informed structured knowledge into model-specific object detection. It transforms attribute and hierarchical knowledge into semantic guidance, enabling a more interpretable and expert-like recognition process. 
Specifically, we design VMAM to associate attribute semantics with visual cues through masked attribute reconstruction, enhancing sensitivity to fine structural differences, and HierVIP to leverage hierarchical knowledge via a visual prototype tree that aligns visual and semantic hierarchies under hierarchical supervised contrastive learning.
Moreover, we curate a fine-grained detection benchmark, PSP, covering 106 ship classes and 30 airplane models, respectively. To the best of our knowledge, PSP provides the most extensive model-specific category coverage among existing fine-grained remote sensing detection datasets, together with consistent hierarchical taxonomies and attribute knowledge.
Comprehensive benchmarking on PSP shows that existing fine-grained detection methods struggle to maintain their advantages when confronted with dense model-specific category spaces and subtle inter-category differences.
ExpertDet embeds structured expert knowledge into the learning process, achieving the best overall performance across datasets and hierarchy levels.

\section*{Acknowledgment}
The authors would like to thank the providers of the open-source datasets and \textit{Google Earth} for the accessibility of remote sensing imagery. We also acknowledge the use of various open-source intelligence resources and technical encyclopedias that facilitate the model-specific annotation of fine-grained ship and aircraft classes.


\bibliographystyle{IEEEtran}
\bibliography{ref}

\end{document}

%% file: secs/intro.tex
\section{Introduction}
\label{sec:intro}

\begin{figure*}[!t]
    \centering
    \includegraphics[width=0.99\linewidth]{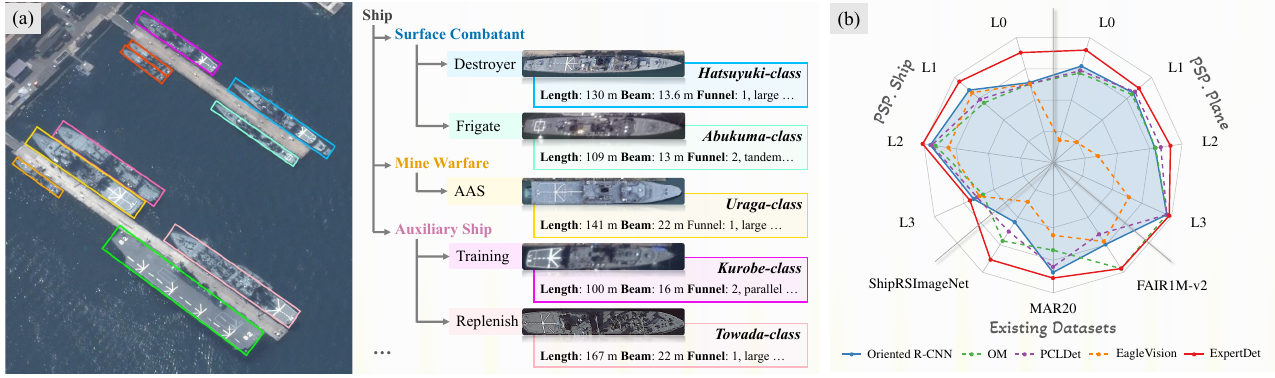}
    \caption{Illustration of benchmark characteristics and performance comparison. (a) A typical image in PSP consisting of many instances from multiple model-specific ship categories; (b) Comparison of ExpertDet with representative fine-grained detection methods on PSP at different hierarchy levels, where L0 denotes the finest model-specific category, as well as on existing fine-grained detection datasets, including MAR20~\cite{wenqi2024mar20}, ShipRSImageNet~\cite{zhang2021shiprsimagenet}, and FAIR1M-v2~\cite{sun2022fair1m}. While existing fine-grained methods show certain effectiveness on benchmarks with coarser category granularity or limited category numbers, their advantages become less consistent on the more challenging PSP benchmark, which contains more model-specific categories with subtler inter-category differences. 
    In comparison, ExpertDet achieves consistent improvements across benchmarks and hierarchy levels. }
  \label{fig:fig1}
  \vspace{-4mm}
\end{figure*}

Aerial object detection~\cite{xia2018dota,li2020dior,ding2021object}, which aims to localize and categorize objects from an overhead perspective, plays a crucial role in a wide range of applications such as urban planning, public security, and surveillance. 
Numerous detection methods~\cite{han2021align,yang2021r3det,xu2021gliding,xie2021oriented,han2021redet} have been proposed, showing promising results for coarse-grained object categories (e.g., airplanes, vehicles, and ships).
However, real-world applications increasingly demand fine-grained, model-specific identification to support higher-level decision-making. 
In contrast to coarse-grained categories, model-specific categories are far more numerous, with distinctions that are often subtle and confined to localized structural details, and typically associated with limited instances and pronounced long-tail distributions~\cite{liu2016hrsc,chen2020fgsd,sun2022fair1m,zhang2021shiprsimagenet,wenqi2024mar20}.
Existing methods largely inherit the paradigm of coarse-grained object detection, predominantly relying on discrete category labels as supervision~\cite{zeng2022instance,li2023petdet,ouyang2023pcldet,zhu2024enhancing}, with the expectation that discriminative visual features can be automatically learned from the data, leading to significant performance degradation in model-specific aerial object detection, as shown in Fig.~\ref{fig:fig1}.

Instead of following a bottom-up paradigm,
human experts rely on domain knowledge to structure the recognition process, aligning prior knowledge with visual observations to distinguish visually similar categories.
Generally, experts exploit two complementary forms of knowledge:
\begin{itemize}
    \item \textbf{Attributes}, which explicitly describe the distinctive structural characteristics of objects. Even when two categories, such as the Boeing 777 and Boeing 787, exhibit highly similar global appearances, experts can distinguish them by recalling key attributes like wingspan or fuselage length. Such attribute knowledge enables systematic discrimination of subtle structural differences across closely related categories. 
    \item \textbf{Hierarchies}, which encode semantic relationships among object categories and provide a structured pathway from coarse-grained to fine-grained recognition. This hierarchical organization can effectively narrow the decision space, reduce confusion among similar categories, and progressively guide attention toward increasingly discriminative semantic levels.
\end{itemize}

Motivated by this, we move beyond forcibly learning model-specific appearances from category labels and embed structured expert knowledge into the learning process to facilitate more targeted fine-grained discrimination.
We present \textit{ExpertDet}, a fine-grained aerial object detection scheme that transforms attributes and hierarchies into semantic guidance, enabling the detector to perceive subtle attribute-level differences while maintaining semantic consistency across hierarchical levels.
Specifically, we introduce \textit{Vision-aware Masked Attribute Modeling} (VMAM), which leverages attribute knowledge to guide the perception of subtle structural cues.
By randomly masking a subset of attribute words and reconstructing them from visual features, VMAM enables the model to associate semantic attributes with their visual counterparts, thereby focusing on attribute-indicated discriminative details.
We further propose \textit{Hierarchical Visual Instance Promotion} (HierVIP), which incorporates hierarchical knowledge to structure visual representations through a visual prototype tree, optimized with a \textit{Hierarchical Supervised Contrastive} (HSC) loss to impose taxonomy-aware constraints on the representation space, preserving semantic continuity between hierarchical levels while promoting discrimination among sibling categories.


Currently, some public datasets~\cite{liu2016hrsc,lam2018xview,chen2020fgsd,zhang2021shiprsimagenet,sun2022fair1m,wenqi2024mar20} have been created to train and evaluate fine-grained object detection methods. 
However, the absence of large-scale fine-grained object detection benchmarks remains a major obstacle to the development of this field.
Notably, ``large-scale'' refers to extensive coverage at the model-specific category level, which is intrinsic to real-world settings characterized by numerous object models.
In this paper, we construct a fine-grained aerial object detection benchmark for \underline{P}recise recognition of model-specific \underline{S}hips and \underline{P}lanes, \textit{PSP}.
It contains 106 ship classes and 30 airplane models, constituting the largest publicly available collection of model-specific categories for aerial object detection to date, with challenges that are well aligned with the long-tailed distributions encountered in real-world applications. Besides, each model-specific category is annotated with explicit hierarchical taxonomies and structure-related attribute knowledge.

In experiments, we integrate ExpertDet into existing state-of-the-art (SOTA) object detectors, including ATSS~\cite{zhang2020atss}, ReDet~\cite{han2021redet}, and Oriented R-CNN~\cite{xie2021oriented}.
Consistent and stable performance gains are observed across all detectors, demonstrating that ExpertDet can effectively leverage expert knowledge to enhance fine-grained discriminative capability.
And we benchmark ExpertDet against a wide range of state-of-the-art object detection methods, spanning both generic object detectors and fine-grained approaches, on the PSP benchmark.
As shown in Fig.~\ref{fig:fig1}, existing fine-grained detection methods, while remaining competitive on benchmarks with coarser category granularity or limited category numbers, struggle to learn stable category representations and under-perform their base detectors on PSP, which contains more model-specific categories with more densely crowded decision boundaries.
ExpertDet goes beyond discrete category supervision by using expert knowledge as semantic guidance to capture fine-grained structural differences, achieving the best overall performance on both existing fine-grained datasets and the more challenging PSP benchmark.


To summarize, our main contributions are three-fold:
\begin{itemize}
    \item We present ExpertDet, a scheme that incorporates expert-informed structured knowledge into fine-grained aerial object detection. Specifically, it transforms attribute and hierarchical knowledge into semantic guidance, 
    effectively improving detection accuracy across multiple fine-grained hierarchy levels.
    \item We design VMAM to link attribute semantics with visual cues through masked attribute reconstruction, enhancing sensitivity to fine structural differences, and HierVIP to leverage hierarchical knowledge via a visual prototype tree that organizes the visual representation space for enhanced inter-class separability.
    \item  To the best of our knowledge, the proposed PSP benchmark provides the most extensive coverage of model-specific categories among existing fine-grained aerial object detection datasets, with each model annotated with structure-related attribute knowledge and organized into explicit hierarchical levels.
\end{itemize}

%% file: secs/related.tex
\section{Related Work}
\label{sec:related}

\begin{table}[!t]
\caption{Comparison of PSP with existing publicly available fine-grained object detection datasets for aircraft and ship models. 
\#Models denotes the number of model-specific categories included in each dataset, 
and \#Instances indicates the number of annotated instances at the model level. Attr. is short for structured attribute annotations.}
\label{tab:data_comp}
\centering
\setlength{\tabcolsep}{3.7pt}
\begin{tabular}{lcccccc} 
\toprule
Dataset & Granularity & \#Models & \#Instances & Attr. & Year \\
\midrule
\textit{plane} &  &  &  &  & \\
xView~\cite{lam2018xview} & Sub. & 0 & 0 & \ding{55} & 2018 \\
FAIR1M~\cite{sun2022fair1m} & Model & 10 & 29,447 & \ding{55} & 2022 \\
MAR20~\cite{wenqi2024mar20} & Model & 20 & 22,341 & \ding{55} & 2024 \\
\rowcolor{gray!20}
PSP & Model  & \textbf{30} & \textbf{51,788} & \ding{51} & 2026 \\ 
\midrule
\textit{ship} &  &  &  &  & \\
HRSC2016~\cite{liu2016hrsc} & Mixed & 9 & 1,482 & \ding{55} & 2016 \\
xView~\cite{lam2018xview} & Sub. & 0 & 0 & \ding{55} & 2018 \\
FGSD~\cite{chen2020fgsd} & Mixed & 26 & 2,620 & \ding{55} & 2020 \\
ShipRSImageNet~\cite{zhang2021shiprsimagenet} & Mixed & 21 & 2,030 & \ding{55} & 2021 \\
FAIR1M~\cite{sun2022fair1m} & Sub. & 0 & 0 & \ding{55} & 2022 \\
MCSD~\cite{guo2023fine} & Sub. & 0 & 0 & \ding{55} & 2023 \\
ORSISOD~\cite{li2025appearance} & Sub. & 0 & 0 & \ding{55} & 2025 \\
\rowcolor{gray!20}
PSP & Model & \textbf{106} & \textbf{5,214} & \ding{51} & 2026 \\
\bottomrule
\end{tabular}
\vspace{-2mm}
\end{table}

\subsection{Fine-Grained Object Detection Methods}
The primary challenge in fine-grained object detection lies in capturing subtle visual cues that distinguish highly similar sub-categories. 
Early approaches address this problem by leveraging auxiliary supervision, such as part annotations, to guide the localization of discriminative object regions. 
For example, Part-based R-CNN~\cite{zhang2014part} introduces dedicated part detectors to explicitly localize informative object components. 
However, the reliance on dense manual annotations limits the scalability of such methods.

To alleviate the need for part-level supervision, subsequent studies focus on enhancing discriminative representation learning. 
Multi-scale feature fusion and contextual modeling have been explored to capture fine-grained structural differences across object categories~\cite{wang2021oriented,song2023fine,li2023petdet,xu2025low,bi2025fghdet}. 
Nevertheless, contextual aggregation may dilute subtle discriminative cues. 
Recently, large vision-language models (LVLMs) have been introduced to enhance semantic understanding in fine-grained detection~\cite{jiang2025eaglevision,ma2025fine}. 
For instance, EagleVision~\cite{jiang2025eaglevision} leverages LVLM-generated instance descriptions to refine visual representations. 
However, since these descriptions are derived mainly from visual content, subtle semantic distinctions may not always be explicitly captured.

Another line of research focuses on optimizing the structure of the feature space to improve inter-class separability. 
Instance-level contrastive objectives~\cite{zeng2022instance,zhu2024enhancing} have been explored to enhance representation discrimination, while prototype-based learning~\cite{ouyang2023pcldet} aggregates information across samples to stabilize feature learning. 
More recently, hierarchical contrastive learning frameworks~\cite{zhang2022hierarchical,ma2025fine} incorporate taxonomy-aware supervision to align representation learning with category hierarchies. 
However, they aggregate contrastive penalties across multiple taxonomic levels, which introduces cross-level semantic interference and may lead to unstable optimization.

\subsection{Fine-grained object detection datasets}
Fine-grained object detection relies on benchmarks with sufficient category granularity to evaluate discrimination among visually similar sub-categories.
An overview of publicly available fine-grained object detection datasets for aircraft and ship models is summarized in Tab.~\ref{tab:data_comp}.
Early benchmarks such as xView~\cite{lam2018xview} make pioneering contributions by introducing a 60-class hierarchical dataset, in which aircraft are subdivided into subcategories such as ``Small Aircraft'' and ``Cargo Plane''.
However, with the increasing demand for model-specific recognition, such taxonomies are insufficient for specialized applications.
FAIR1M~\cite{sun2022fair1m} further extends the taxonomic granularity of xView by providing 37 fine-grained categories, including specific aircraft models such as ``Boeing 737'' and ``Airbus A320''.
Nevertheless, the number of model-specific categories remains limited, and taxonomic inconsistencies persist.
For example, the generic category ``other-airplane'' is placed at the same hierarchical level as model-specific classes.
MAR20~\cite{wenqi2024mar20} includes a broader range of aircraft models and provides annotations for 20 specific types. However, these models are treated as isolated categories without explicit parent–child relationships.
Similarly, HRSC2016~\cite{liu2016hrsc}, FGSD~\cite{chen2020fgsd}, and ShipRSImageNet~\cite{zhang2021shiprsimagenet} refine annotations to specific ship models. However, their taxonomic depth remains constrained, and hierarchical inconsistencies are also observed. For example, FGSD places specific ship classes (e.g., Asagiri-class) alongside unresolved generic categories (e.g., Destroyers) at the same hierarchical level, resulting in ambiguous category organization.

%% file: secs/method.tex
\section{Methodology}
\subsection{Overview}
ExpertDet learns to detect fine-grained objects by incorporating structured expert knowledge, \ie, attributes for capturing subtle yet discriminative structural differences, and hierarchies for encoding semantic consistency across granularities, thereby moving beyond conventional single-label supervision.
The core idea is conceptually simple, in which two complementary modules are plugged in to activate these two forms of knowledge: 
(1) Vision-aware Masked Attribute Modeling (VMAM), which aligns attribute semantics with visual structures by reconstructing randomly masked attributes from visual cues, thereby enhancing sensitivity to fine-grained structural differences;
(2) Hierarchical Visual Instance Promotion (HierVIP), which leverages hierarchical relations to build a visual prototype tree, structuring the feature space to enhance inter-class separability.

\begin{figure*}[!t]
  \centering
  \includegraphics[width=0.92\linewidth]{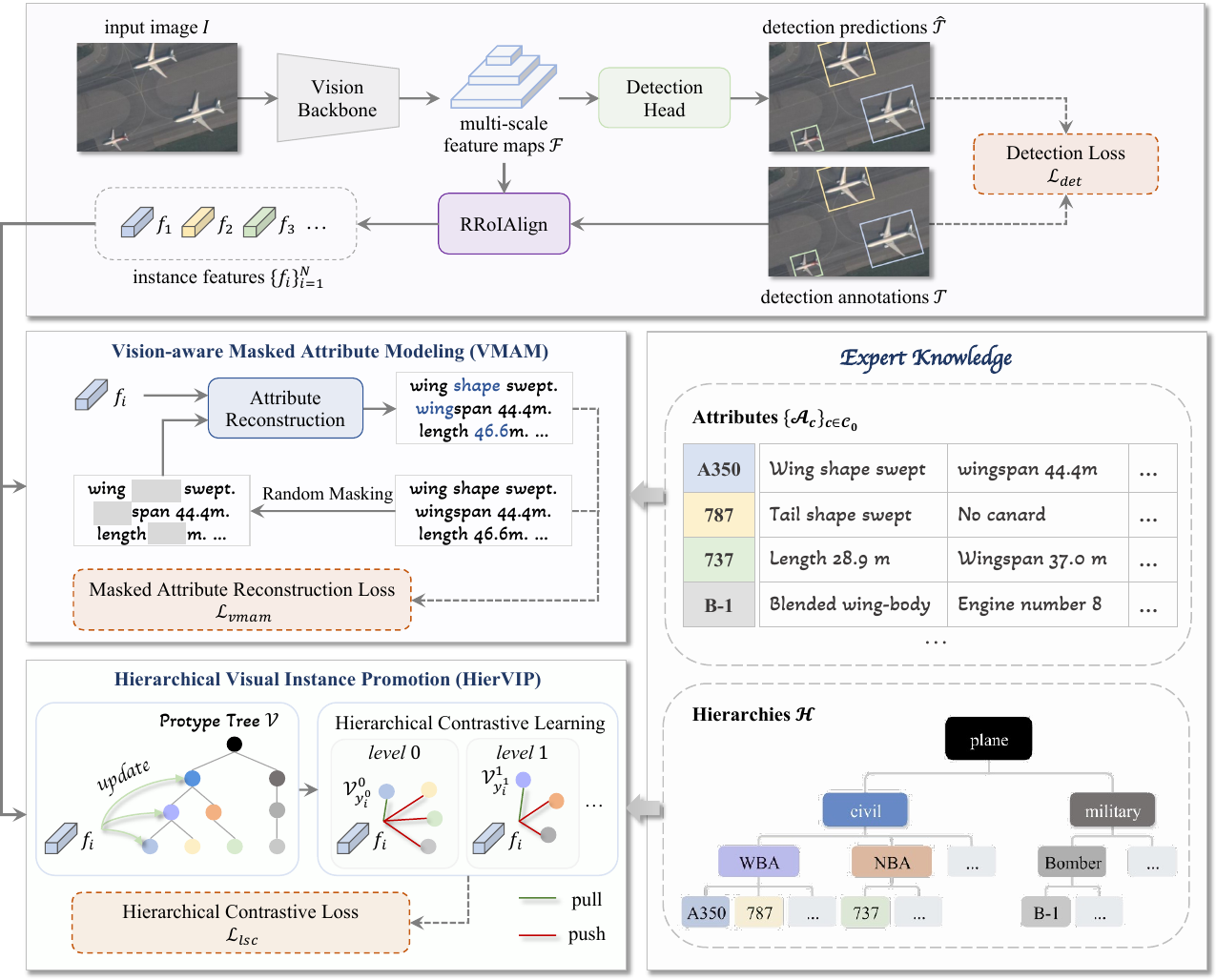}
    \caption{The architecture of the proposed ExpertDet framework.}
  \label{fig:overview}
\vspace{-4mm}
\end{figure*}

The overall architecture of ExpertDet is illustrated in Fig.~\ref{fig:overview}.
To train ExpertDet, besides the standard detection annotations $\mathcal{T}=\{(b_i,y_i^{0})\}_{i=1}^{N}$, where $N$ denotes the number of instances, and $(b_i,y_i^{0})$ denote the bounding box coordinates and model-specific category label of the $i$-th instance, respectively, we further require two forms of expert knowledge for supervision: a hierarchical taxonomy $\mathcal{H}=\{\mathcal{C}_{l}\}_{l=0}^{L-1}$ spanning all model-specific categories, where $L$ is the number of hierarchy levels, $\mathcal{C}_{0}$ denotes the set of model-specific categories and $\mathcal{C}_{l}$ for $l>0$ denote the category sets at progressively coarser levels; and a set of attribute entries $\mathcal{A}_{c}=\{a_{c,k}\}_{k=1}^{K_c}$ for each model-specific category $c\in\mathcal{C}_{0}$, where $K_c$ denotes the number of attribute entries associated with $c$, and $a_{c,k}$ denotes its $k$-th attribute entry.
Accordingly, each instance $t_i\in\mathcal{T}$ is associated with a hierarchy label path $\{y_i^{l}\}_{l=0}^{L-1}$, derived from its model-specific category label $y_i^{0}$ under the taxonomy $\mathcal{H}$, where $y_i^{l}\in\mathcal{C}_{l}$, together with the corresponding set of attribute entries  $\mathcal{A}_{y_i^{0}}$.

In the training stage, an input image $I$ is first fed into the vision backbone to extract multi-scale feature maps $\mathcal{F}$, 
which are then forwarded to the detection network to produce detection predictions 
$\hat{\mathcal{T}}=\{(\hat{b}_i,\hat{y}_i^{0})\}_{i=1}^{\hat{N}}$, where $\hat{N}$ denotes the number of predicted instances, and $\hat{b}_i$ and $\hat{y}_i^{0}$ denote the predicted bounding box and model-specific category label, respectively. 
Based on $\hat{\mathcal{T}}$ and the corresponding detection annotations $\mathcal{T}$, the standard detection loss $\mathcal{L}_{det}$ is computed.
Meanwhile, for each instance $t_i$, we extract its instance-level visual feature $f_i$ from $\mathcal{F}$ according to the ground-truth box coordinates $b_i$ via a Rotated RoI Align layer $\mathrm{RRoIAlign}(\cdot)$: 
\begin{equation}
f_i=\mathrm{RRoIAlign}(\mathcal{F}, b_i).
\end{equation}
The extracted instance-level features are further supervised by the two forms of expert knowledge, \ie, attributes and hierarchies, associated with each instance.
Specifically, VMAM takes $\{f_i\}_{i=1}^{N}$ together with the corresponding sets of attribute entries $\{\mathcal{A}_{y_i^{0}}\}_{i=1}^{N}$ to perform masked attribute reconstruction, yielding the masked attribute reconstruction loss $\mathcal{L}_{vmam}$. 
In parallel, HierVIP incorporates $\{f_i\}_{i=1}^{N}$ to update the visual prototype tree constructed from the taxonomy $\mathcal{H}$, upon which hierarchical contrastive learning is imposed to yield the hierarchical contrastive loss $\mathcal{L}_{hsc}$.
The overall training objective of ExpertDet is defined as:
\begin{equation}
\mathcal{L}=\mathcal{L}_{det}+\lambda\mathcal{L}_{vmam}+\mu\mathcal{L}_{hsc},
\end{equation}
where $\lambda$ and $\mu$ are balancing factors.

During the inference stage, only the standard detection pipeline is retained, while VMAM and HierVIP are discarded, introducing no additional computational overhead.

\subsection{Vision-aware Masked Attribute Modeling}
\label{sec:vmam}
To enable the detector to capture subtle yet discriminative differences in an expert-like manner, 
it is necessary to establish explicit correspondence between attribute descriptions and visual features. 
A straightforward solution is to recover the entire attribute description directly from visual features. 
However, such an open-ended generation objective suffers from an excessively large output space and unstable optimization, 
while also introducing redundant and generic semantics that dilute supervision away from the local discriminative cues critical for distinguishing object models, thereby weakening the precise alignment between attribute semantics and visual features.

Therefore, instead of recovering the entire attribute description, VMAM formulates attribute learning as a masked reconstruction task.
This formulation is also more consistent with expert cognition, 
since experts do not perceive all attributes uniformly, 
but instead first anchor on partially observed cues and then leverage inter-attribute dependencies to identify the remaining model-critical attributes from visual features.
In this way, visual features are encouraged to encode both inter-attribute dependencies and the local structural cues necessary for identifying model-critical attributes.

Specifically, for each instance $t_i$ with model-specific category label $y_i^{0}$, the textual attribute description $s_i$ is formed by concatenating the associated attribute entries $\mathcal{A}_{y_i^{0}}$, which are randomly shuffled to avoid positional bias among attributes.
We first tokenize $s_i$ into an attribute token sequence $\mathbf{z}_i=\{z_{i,j}\}_{j=1}^{M}$, where $M$ is the maximum token length.
Following the masked language modeling (MLM) paradigm~\cite{devlin2019bert}, a subset of tokens is randomly masked with probability $q$. Let $\mathcal{M}_i \subseteq \{1,\ldots,M\}$ denote the set of masked positions for the $i$-th instance, 
and let $\tilde{\mathbf{z}}_i=\{\tilde z_{i,j}\}_{j=1}^{M}$ denote the resulting masked token sequence. 
The masked sequence $\tilde{\mathbf{z}}_i$ is then fed into a language encoder to obtain textual embeddings $E_i$.
The textual embeddings $E_i$ and the corresponding instance-level visual feature $f_i$ are jointly processed by a vision-language encoder, which facilitates cross-modal interaction via self-attention and yields fused multi-modal representations $E_i^{vl}$.
A prediction head is applied to $E_i^{vl}$ to estimate the probability distributions $\mathbf{p}_i=\{p_{i,j}\}_{j=1}^{M}$, where ${p}_{i,j}$ denotes the predicted distribution at the $j$th token position for the $i$th instance. The masked attribute reconstruction loss $\mathcal{L}_{vmam}$ is then defined:
\begin{equation}
    \mathcal{L}_{vmam}=\frac{1}{\sum_{i=1}^{N_m}\!|\mathcal{M}_i|}\!\sum_{i=1}^{N_m}\!\sum_{j\in\mathcal{M}_i}\!\mathrm{CrossEntropy}({p}_{i,j},z_{i,j}),
\end{equation}
where $\mathrm{CrossEntropy}(\cdot)$ is the cross-entropy loss, and $N_m$ denotes the number of instances processed in VMAM to control computational cost.
when $N>N_m$, $N_m$ instances are randomly sampled for training.

\subsection{Hierarchical Visual Instance Promotion}

Fine-grained detection often involves a high density of model-specific categories within a single parent category, giving rise to a highly crowded feature space where stable and discriminative decision boundaries are difficult to learn.
Indiscriminately distancing a specific model-specific category from all others introduces an excessive number of pairwise separation constraints and significantly increases optimization difficulty.
In practice, the feature space should be able to distinguish subtle differences among model-specific categories while preserving reasonable semantic continuity among related ones. Such semantic continuity further benefits sample-scarce categories, whose representations can still be properly constrained in the feature space by the higher-level semantics shared with related categories, thereby facilitating the establishment of more robust fine-grained decision boundaries.
Therefore, HierVIP is designed to structure the feature space according to the hierarchical taxonomy, thereby preserving semantic continuity across hierarchy levels while enhancing discrimination among related model-specific categories.

Specifically, we construct a visual prototype tree $\mathcal{V}=\{\mathcal{V}^{l}\}_{l=0}^{L-1}$ based on the hierarchical taxonomy $\mathcal{H}=\{\mathcal{C}_{l}\}_{l=0}^{L-1}$, 
where $\mathcal{V}^{l}=\{V_c^{l}\mid c\in\mathcal{C}_{l}\}$ denotes the set of prototypes at hierarchy level $l$, and $V_c^{l}$ is the prototype corresponding to category $c\in\mathcal{C}_{l}$.
For each instance $t_i$ belonging to model-specific category $y_i^{0}$, the corresponding prototypes $\{V_{y_i^{l}}^{l}\}_{l=0}^{L-1}$ along its hierarchy label path $\{y_i^{l}\}_{l=0}^{L-1}$ are updated using its visual feature $f_i$ via a momentum-based aggregation strategy:
\begin{gather}
    V_{y_i^{l}}^{l} \leftarrow w V_{y_i^{l}}^{l} + (1-w) f_i, \\
    V_{y_i^{l}}^{l} \leftarrow \frac{V_{y_i^{l}}^{l}}{\|V_{y_i^{l}}^{l}\|_2},
\end{gather}
where $w$ denotes the momentum coefficient.
If a prototype is activated for the first time, it is initialized by directly assigning the instance feature, \ie, $V_{y_i^{l}}^{l} = f_i$.

Since model-specific categories are typically associated with limited instances, 
a large momentum coefficient $w$ may cause prototypes to be overly influenced by noisy instance features, leading to unstable updates, whereas a small $w$ slows down adaptation and results in a mismatch between instance features and outdated prototypes, thereby degrading the effectiveness of representation learning. To address this issue, we adopt an adaptive momentum strategy that dynamically adjusts $w$ according to the feature drift during training. Specifically, the drift between the current prototype and the incoming instance feature is measured as:
\begin{gather}
    d_i = 1 - (V^{l}_{y_i^l} \cdot f_i).
\end{gather}
A larger $d$ indicates a more significant shift in the category representation, suggesting that the prototype should adapt more rapidly. Accordingly, the momentum coefficient is dynamically computed as follows:
\begin{equation}
w =\operatorname{clip}\left(w_{\text{base}} - \alpha d_i,\;w_{\text{low}},\;w_{\text{base}}\right),
\end{equation}
where $\operatorname{clip}(\cdot)$ constrains $w$ to the interval $[w_{\text{low}}, w_{\text{base}}]$, and $\alpha$ controls the sensitivity to feature drift. We empirically set $w_{\text{low}}=0.5$, $w_{\text{base}}=0.8$, and $\alpha=0.2$.

We formulate a hierarchical contrastive objective over the prototype tree to explicitly enforce level-specific discriminative constraints while maintaining taxonomic coherence. 
For each instance $t_i$, we calculate the cosine similarity between its visual feature $f_i$ and all activated prototypes at hierarchy level $l$: 
\begin{equation}
s_{i,c}^{l}=\frac{f_i^{\top}V_c^{l}}{\tau}, \quad c\in\mathcal{C}_{l}^{act},
\end{equation}
where $\tau$ is a temperature parameter set to 0.2, and $\mathcal{C}_{l}^{act} \subseteq \mathcal{C}_{l}$ denotes the subset of categories whose prototypes have been activated at level $l$.
The probability of assigning $t_i$ to its ground-truth category $y_i^{l}$ at level $l$ is defined as:
\begin{equation}
P(y_i^{l}\mid t_i) = \frac{\exp(s_{i,y_i^{l}}^{l})}
{\sum_{c\in\mathcal{C}_{l}^{act}}\exp(s_{i,c}^{l})},
\end{equation}
Only activated prototypes are included in the loss computation to ensure stable optimization.
The hierarchical contrastive loss is then defined as the weighted negative log-likelihood over all informative hierarchy levels:
\begin{equation}
\mathcal{L}_{hsc}
=
-\frac{1}{N}
\sum_{i=1}^{N}
\frac{
\sum_{l=0}^{L-1}\gamma_l\log P(y_i^{l}\mid t_i)
}{
\sum_{l=0}^{L-1}\gamma_l
},
\end{equation}
where $\gamma_l$ denotes the level-specific weight. We set $\gamma_l=L-l-1$, assigning larger weights to finer levels while excluding the root level.

%% file: secs/dataset.tex
\section{Model-specific Object Detection Dataset}

To facilitate model-specific aerial object detection, we focus on two representative object categories in aerial imagery, ship and aircraft, and construct a fine-grained object detection benchmark, \textit{PSP}. Unlike existing fine-grained datasets that often mix specific classes (e.g., Asagiri-class) with generic, unresolved sub-categories (e.g., Destroyer) at the same taxonomic level, our datasets introduce a carefully curated hierarchical taxonomy that organizes object models in a consistent and fine-grained manner. Within this taxonomy, \textit{PSP.Ship} contains 106 ship models and \textit{PSP.Plane} includes 30 airplane models, providing the most extensive coverage of model-specific categories among existing fine-grained aerial object detection datasets to date. The details of \textit{PSP.Ship} and \textit{PSP.Plane} are described below.

\subsection{PSP.Ship}
\begin{figure*}
  \centering
  \includegraphics[width=0.99\linewidth]{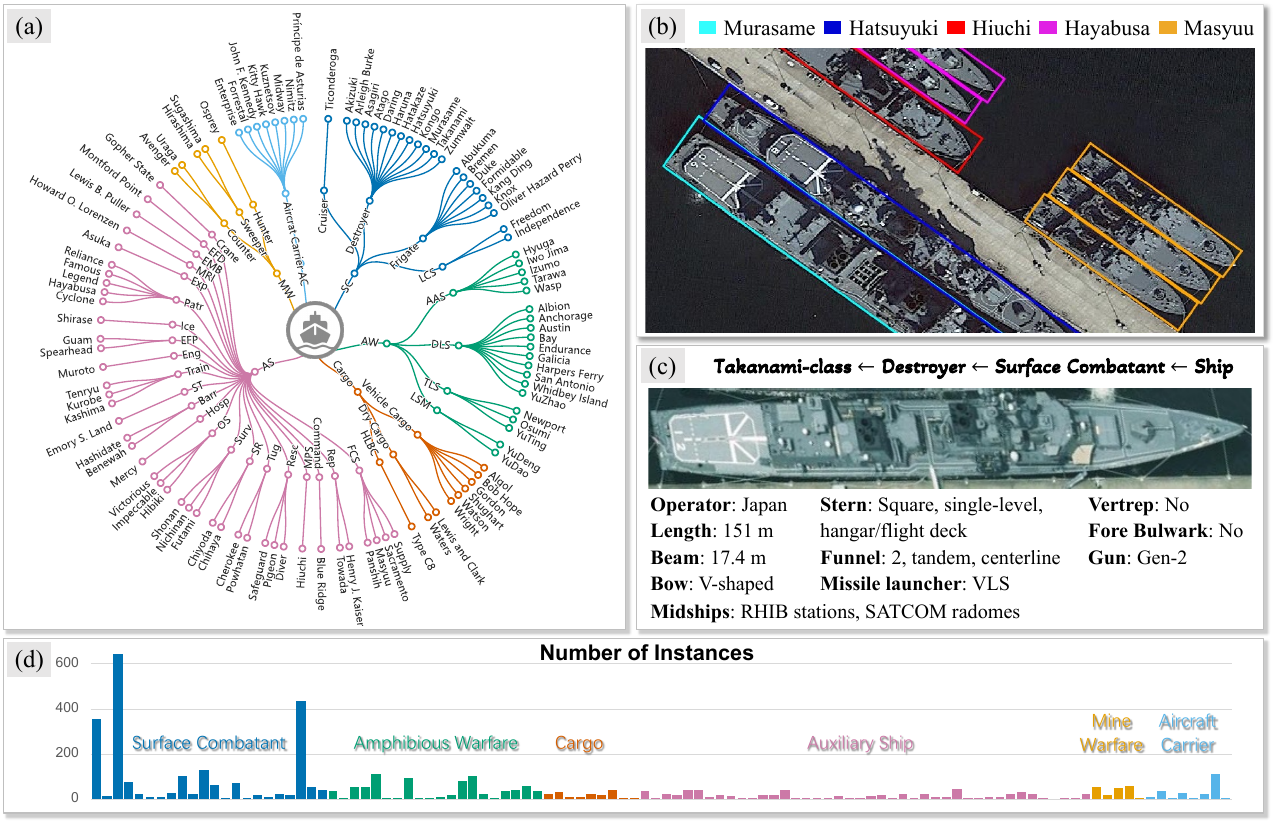}
  \caption{Illustration of PSP.Ship. (a) Hierarchical taxonomy of the 106 ship models in PSP.Ship; (b) An example image from PSP.Ship; (c) A representative instance from PSP.Ship annotated with 11 attributes; (d) Number of instances per ship model. }
  \label{fig:ship_ov}
\vspace{-4mm}
\end{figure*}

The images in \textit{PSP.Ship} are collected from Google Earth imagery and selected images in existing datasets, including ShipRSImageNet~\cite{zhang2021shiprsimagenet}, MCSD~\cite{guo2023fine}, and DOTA-v2~\cite{ding2021object}, with spatial resolutions ranging from 0.1\,m to 2\,m.
For the Google Earth imagery, ship instances are annotated with rotated bounding boxes and model-specific labels. 
For the selected images in existing datasets, the annotations are derived from the original annotations.
Notably, instances in ShipRSImageNet are partially annotated with specific ship models, while others are labeled with generic categories such as \textit{other destroyer}.
Meanwhile, ship instances in MCSD are annotated with subcategory labels (e.g., \textit{destroyer}), whereas instances in DOTA-v2 are labeled only at the coarse category level (i.e., \textit{ship}).
We therefore refine these annotations by assigning model-specific ship labels to instances from ShipRSImageNet with unresolved labels, as well as those from MCSD and DOTA-v2.
In addition, we observe that a portion of the bounding box annotations are either horizontal or fail to tightly enclose the objects. We therefore manually correct these annotations into more accurate rotated bounding boxes.
Overall, \textit{PSP.Ship} contains 1,953 images with 5,214 ship instances covering 106 ship models, as illustrated in Fig.~\ref{fig:ship_ov}. 
The dataset is randomly split into training and test sets at a ratio of 4:1. The training set contains 1,563 images with 4,181 ship instances spanning 105 ship models, while the test set contains 390 images with 1,033 ship instances spanning 91 ship models.


Instead of treating each ship model as an entirely independent category, we organize the 106 ship models into a four-level hierarchical taxonomy following publicly available ship classification schemes\footnote{\url{https://en.wikipedia.org/wiki/Naval_ship}}. 
The first level corresponds to the root category \textit{ship}. 
The second level groups ships according to their operational roles, comprising six categories: \textit{surface combatants} (SC), \textit{amphibious warfare ships} (AW), \textit{auxiliary ships} (AS), \textit{aircraft carriers} (AC), \textit{cargo vessels} (Cargo), and \textit{mine warfare vessels} (MW).
The third level corresponds to more specific ship types under each operational role. For example, \textit{surface combatants} are further divided into four ship types: \textit{Cruiser}, \textit{Destroyer}, \textit{Frigate}, and \textit{Littoral Combat Ship}.
The fourth level corresponds to specific ship classes, which constitute the model-specific categories in the dataset, such as \textit{Akizuki-class}, \textit{Arleigh Burke-class}, and \textit{Asagiri-class}.

Furthermore, for each ship class, we annotate a set of attributes derived from publicly available information in Wikipedia. These attributes capture visual cues from three perspectives: absolute size cues, including \textit{overall length} and \textit{beam}; stylistic semantic cues, represented by the \textit{operator}; and structural detail cues, including \textit{bow}, \textit{stern}, \textit{midships structure}, \textit{funnel}, \textit{fore bulwark}, \textit{vertical replenishment point}, \textit{gun type}, and \textit{missile launcher type}.
In total, 11 attributes are annotated, whose values include numerical measurements (e.g., the value of \textit{beam} is a numerical quantity), binary indicators (e.g., \textit{vertical replenishment point} indicates the presence or absence of the structure), categorical labels (e.g., \textit{missile launcher type} takes one value from a predefined set of categories), and textual descriptors (e.g., \textit{midships structure} is described using textual phrases that characterize the configuration).

\subsection{PSP.Plane}
\begin{figure}
  \centering
  \includegraphics[width=0.99\linewidth]{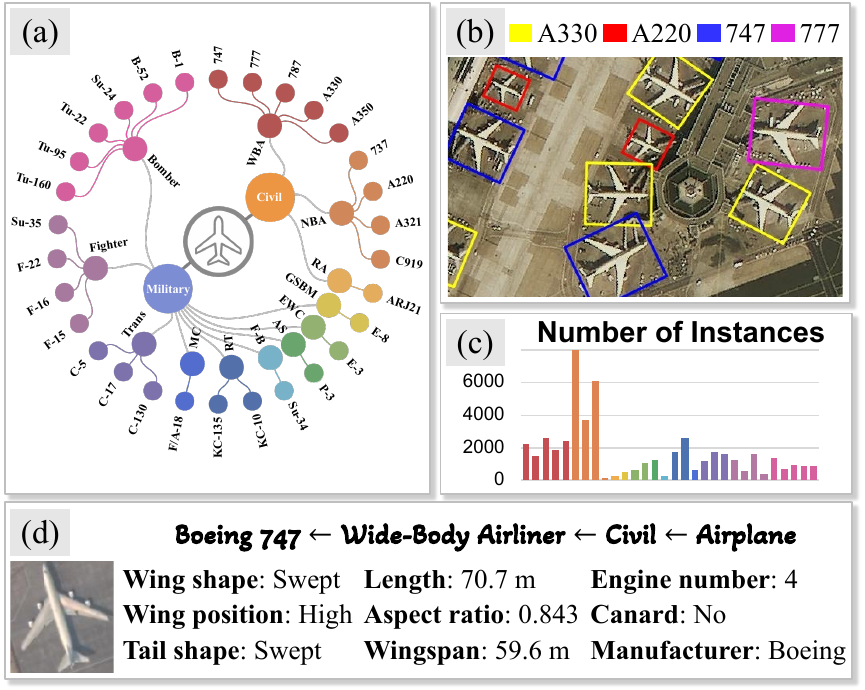}
  \caption{Illustration of PSP.Plane. (a) Hierarchical taxonomy of the 30 plane models in PSP.Plane; (b) An example image from PSP.Plane; (c) Number of instances per plane model; (d) A representative instance from PSP.Plane annotated with 9 attributes. }
  \label{fig:plane_ov}
\vspace{-4mm}
\end{figure}

The images in \textit{PSP.Plane} are collected from existing datasets, MAR20~\cite{wenqi2024mar20} and FAIR1M-v2~\cite{sun2022fair1m}, with spatial resolutions ranging from 0.3\,m to 1.1\,m. 
We retain the original category annotations and directly adopt the official training and validation splits of MAR20 and FAIR1M-v2, which are then merged into a unified dataset. The resulting \textit{PSP.Plane} dataset contains 6,192 training images with 28,216 instances and 4,567 validation images with 23,572 instances, while both splits cover 30 airplane model categories.

Similarly to \textit{PSP.Ship}, we organize the 30 airplane models into a four-level hierarchical taxonomy referring to the publicly available airplane classification 
schemes\footnote{\url{https://en.wikipedia.org/wiki/Military_aircraft}}\textsuperscript{,}\footnote{\url{https://en.wikipedia.org/wiki/Airliner}}.
The first level corresponds to the root category \textit{plane}. 
The second level groups planes by usage into \textit{military} (M) and \textit{civil} (C). 
The third level further refines each branch into functional plane types. 
For instance, the civil branch includes \textit{wide-body airliner}, \textit{narrow-body airliner}, and \textit{regional airliner}. 
The fourth level corresponds to specific plane models, such as \textit{Airbus A330}, \textit{Boeing 747}, and \textit{COMAC C919}.

For each airplane model, we annotate 9 attributes that cover the same aspects as those in \textit{PSP.Ship}, i.e., absolute size cues, stylistic semantic cues, and structural detail cues. Specifically, the absolute size cues include \textit{wingspan}, \textit{length}, and \textit{aspect ratio}; the stylistic semantic cue is represented by the \textit{manufacturer}; and the structural detail cues include \textit{wing shape}, \textit{wing position}, \textit{engine number}, \textit{canard}, and \textit{tail shape}.

\subsection{Properties of PSP}

\textbf{Numerous model-specific Categories.} 
\textit{PSP.Ship} contains 106 ship models and \textit{PSP.Plane} includes 30 aircraft models, substantially expanding beyond ShipRSImageNet (26 ship models) and MAR20 (20 aircraft models), the existing benchmarks with the broadest model coverage. Moreover, as these categories belong to the same coarse category, visual differences between individual models are often subtle and confined to localized structural details. Consequently, the enlarged category space becomes highly crowded, substantially aggravating inter-category confusion and imposing greater demands on detectors to learn stable and discriminative fine-grained representations.

\textbf{Category Imbalance.} 
\textit{PSP} exhibits a pronounced long-tail distribution, where a few categories contain abundant instances while many others are rarely observed. In particular, in \textit{PSP.Ship}, the top three categories account for over 30\% of all instances. This imbalance mirrors real-world category distributions, while calling for detectors that avoid overfitting to categories with abundant instances and preserve robust performance on rare categories.

\textbf{Sample Scarcity.} 
Beyond the relative category imbalance, \textit{PSP.Ship} is further characterized by an absolute scarcity of samples across the majority of its specific categories. Specifically, more than 70 categories in \textit{PSP.Ship} contain fewer than 40 annotated instances. This scarcity stems from the intrinsic rarity of many models in real-world scenes, while requiring detectors to learn discriminative representations under extremely limited supervision.

%% file: secs/exp.tex
\section{Experiments}
\subsection{Experimental Setup}
\textbf{Datasets}.
In addition to experiments on PSP.Ship and PSP.Plane, we conduct experiments on three more fine-grained object detection datasets, i.e, MAR20~\cite{wenqi2024mar20}, ShipRSImageNet~\cite{zhang2021shiprsimagenet}, and FAIR1M-v2~\cite{sun2022fair1m}, evaluate the effectiveness of the proposed method.

\textbf{Implementation Details}.
We conduct all the experiments on an NVIDIA RTX 3090 GPU with a batch size of 4 by default.
ExpertDet adopts Oriented R-CNN~\cite{xie2021oriented} with Swin-T~\cite{liu2021swin} backbone pretrained on ImageNet~\cite{deng2009imagenet} as the baseline detector unless specified otherwise.
The language encoder in VMAM is implemented using BERT~\cite{devlin2019bert}. The maximum instance number $N_m$ is set to 8. The mask probability $q$ is set to 0.1 for PSP.Ship and 0.9 for the other datasets. 
The weighting factors $\lambda$ and $\mu$ are set to $0.2$ and $0.001$, respectively.
We only use random flipping as data augmentation for all experiments.
For experiments on PSP.Ship, PSP.Plane and MAR20, images are resized such that their shorter side is 800 pixels, whereas for experiments on ShipRSImageNet and FAIR1M-v2, images are resized such that their shorter side is 1024 pixels.
All models are trained for 24 epochs using the Adam optimizer with a learning rate of $5\times 10^{-5}$ and a weight decay of 0.05.

\textbf{Evaluation metric}.
Unless otherwise specified, the performance of each category refers to the average detection accuracy of all model-specific categories under it, not to the detection accuracy of the category itself. For example, the performance of \textit{Ship} is computed as the average detection accuracy over all ship models, whereas the performance of \textit{Destroyer} is computed as the average detection accuracy over the 12 ship models within this category. We adopt the mean Average Precision (mAP) at an IoU threshold of 0.5 as the evaluation metric.

\begin{table*}[!bt]
\setlength{\tabcolsep}{4.2pt}
\centering
\caption{Experimental results of general-purpose, fine-grained, and open-vocabulary object detection methods on PSP.Ship, covering one-stage, two-stage, and Transformer-based architectures.  }
\label{tab:overall_ship}
\begin{threeparttable}
\begin{tabular}{l|c|c|c|c|cccccc|cc}
\toprule
Method & Publication & Backbone & Sch. & Ship & SC & AW & CG & AS & MW & AC &\#Params.&GFLOPs \\
\midrule
\rowcolor{gray!20}
\multicolumn{1}{l}{\textit{One-stage}} & \multicolumn{1}{c}{}&\multicolumn{1}{c}{} & \multicolumn{1}{c}{}& \multicolumn{1}{c}{}& & & & & & \multicolumn{1}{c}{}& & \\
R. RetinaNet~\cite{lin2020retinanet} & TPAMI 2020 & Swin-T & 2$\times$ & 24.9 & 18.3 & 32.7 & 14.6 & 19.7 & 5.9 & 31.9 & 37.2 & 161.7 \\
R. FCOS~\cite{tian2019fcos} & ICCV 2019 & Swin-T & 2$\times$ & 59.0 & 47.4 & 59.9 & 48.1 & 46.0 & 51.5 & 66.2 & 35.1 & 131.6 \\
R. ATSS~\cite{zhang2020atss} & CVPR 2020 & Swin-T & 2$\times$ & 61.0 & 49.6 & 57.9 & 44.7 & 49.3 & 44.7 & 77.5 & 35.3 & 134.3 \\
R$^3$Det~\cite{yang2021r3det} & AAAI 2021 & Swin-T & 2$\times$ & 32.1 & 33.6 & 33.8 & 20.3 & 20.2 & 22.7 & 45.3 & 42.6 & 237.7 \\
S$^2$A-Net~\cite{han2021align} & TGRS 2021 & Swin-T & 2$\times$ & 45.5 & 45.0 & 49.1 & 38.7 & 28.9 & 21.6 & 63.1 & 37.7 & 128.1 \\
R. GLIP~\cite{li2022glip} & CVPR 2022 & Swin-T & 2$\times$ & 57.4 & 59.4 & 56.5 & 35.4 & 41.3 & 32.3 & 71.4 & 221.2 & 150.0 \\
OM (R. FCOS)~\cite{zhu2024enhancing} & IGARSS 2025 & Swin-T & 2$\times$ & 60.8 \scriptsize{\mygreen{(+1.8)}} & 56.3 & 56.0 & 48.2 & 44.2 & 45.2 & 81.8 & 35.3 & 134.4 \\
\midrule
\rowcolor{gray!20}
\multicolumn{1}{l}{\textit{Transformer-based}} & \multicolumn{1}{c}{}&\multicolumn{1}{c}{} & \multicolumn{1}{c}{}& \multicolumn{1}{c}{}& & & & & & \multicolumn{1}{c}{}& &\\
R. DeformableDETR~\cite{zhu2021deformable} & ICLR 2021 & Swin-T & 50e & 30.1 & 30.5 & 36.2 & 18.6 & 20.9 & 15.4 & 29.0 & 40.9 & 136.0 \\
RHINO~\cite{lee2024hausdorff} & WACV 2025 & Swin-T & 2$\times$ & 63.8 & 63.8 & 55.5 & 46.5 & 50.7 & 36.4 & 68.5 & 51.0 & 193.0 \\
R. Grounding DINO~\cite{liu2024grounding} & ECCV 2024 & Swin-T & 2$\times$ & 58.6 & 55.3 & 50.7 & 50.0 & 44.3 & 40.1 & 72.7 & 164.9 & 152.0 \\
\midrule
\rowcolor{gray!20}
\multicolumn{1}{l}{\textit{Two-stage}} & \multicolumn{1}{c}{}&\multicolumn{1}{c}{} & \multicolumn{1}{c}{}& \multicolumn{1}{c}{}& & & & & & \multicolumn{1}{c}{}& &\\
R. Faster R-CNN~\cite{ren2016faster} & TPAMI 2016 & Swin-T & 2$\times$ & 61.9 & 57.6 & 56.9 & 51.9 & 48.7 & 21.0 & 75.9 & 44.4 & 138.0 \\
Oriented R-CNN~\cite{xie2021oriented} & ICCV 2021 & Swin-T & 2$\times$ & \underline{79.3} & 79.8 & \underline{69.6} & \textbf{68.6} & \underline{58.9} & 41.8 & \underline{93.7} & 44.5 & 138.2 \\
ReDet~\cite{han2021redet} & CVPR 2021 & ReR50 & 2$\times$ & 68.5 & 69.7 & 67.0 & 56.9 & 50.4 & 50.1 & 59.7 & 30.8 & 60.5 \\
Gliding Vertex~\cite{xu2021gliding} & TPAMI 2021 & Swin-T & 2$\times$ & 62.1 & 58.0 & 50.7 & 57.2 & 48.0 & 50.2 & 71.4 & 44.5 & 138.2 \\
EQLv2~\cite{tan2023eql} (Oriented R-CNN) & TPAMI 2023 & Swin-T & 2$\times$ & 60.3 \scriptsize{\myred{(-19.0)}} & 62.2 & 57.2 & 41.3 & 51.5 & 26.8 & 37.1 & 44.5 & 138.1 \\
LogN~\cite{zhao2024logit} (Oriented R-CNN) & IJCV 2024 & Swin-T & 2$\times$ & 71.9 \scriptsize{\myred{(-7.4)}} & 70.5 & \textbf{70.0} & 62.5 & 50.5 & 48.9 & 83.3 & 44.5 & 138.2 \\
R. HiCLPL~\cite{zhang2022hierarchical} (R. Faster R-CNN) & ACMMM 2022 & Swin-T & 2$\times$ & 16.1 \scriptsize{\myred{(-45.8)}} & 9.7 & 21.3 & 12.9 & 10.2 & 6.5 & 34.3 & 46.7 & 138.2 \\
ISCL~\cite{zeng2022instance} (Oriented R-CNN) & TGRS 2022 & Swin-T & 2$\times$ & 76.4 \scriptsize{\myred{(-2.9)}}& 80.3 & 65.9 & 71.1 & 53.8 & 44.3 & 89.8 & 58.2 & 152.7 \\
PETDet~\cite{li2023petdet} (Oriented R-CNN) & TGRS 2023 & Swin-T & 2$\times$ & 76.7 \scriptsize{\myred{(-2.6)}}& 79.4 & 67.4 & 55.6 & 57.6 & \textbf{61.7} & 79.3 & 46.2 & 133.0 \\
PCLDet~\cite{ouyang2023pcldet} (Oriented R-CNN) & TGRS 2023 & Swin-T & 2$\times$ & 78.4 \scriptsize{\myred{(-0.9)}}& \underline{80.4} & 66.0 & 63.7 & 58.0 & 58.5 & 89.8 & 46.0 & 139.8 \\
OM~\cite{zhu2024enhancing} (Oriented R-CNN) & IGARSS 2025 & Swin-T & 2$\times$ & 77.9 \scriptsize{\myred{(-1.4)}} & 79.2 & 67.7 & 62.8 & 56.8 & 55.3 & 93.1 & 45.6 & 138.2 \\
EagleVision~\cite{jiang2025eaglevision} (Oriented R-CNN) & arXiv 2025 & Swin-T & 3$\times$ & 64.5 \scriptsize{\myred{(-14.8)}}& 61.1 & 63.4 & 59.3 & 46.0 & 37.4 & 75.7 & 669.0 & 142.0 \\
\midrule
\rowcolor{gray!20}
\multicolumn{1}{l}{\textit{Ours}} & \multicolumn{1}{c}{}&\multicolumn{1}{c}{} & \multicolumn{1}{c}{}& \multicolumn{1}{c}{}& & & & & & \multicolumn{1}{c}{}& &\\
\rowcolor{blue!5}
ExpertDet (R. ATSS) & - & Swin-T & 2$\times$ & 63.0 \scriptsize{\mygreen{(+2.0)}} & 54.0 & 55.8 & 48.4 & 50.4 & 54.5 & 75.9 & 34.2 & 134.3 \\
\rowcolor{blue!5}
ExpertDet (ReDet) & - & ReR50 & 2$\times$ & 70.9 \scriptsize{\mygreen{(+2.4)}} & 69.2 & 67.1 & 60.4 & 52.3 & 55.2 & 70.4 & 31.0 & 60.5 \\
\rowcolor{blue!5}
ExpertDet (Oriented R-CNN) & - & Swin-T & 2$\times$ & \textbf{82.5} \scriptsize{\mygreen{(+3.2)}}& \textbf{81.1} & 68.5 & \underline{68.5} & \textbf{63.4} & \underline{60.3} & \textbf{93.9} & 44.6 & 138.2 \\
\bottomrule
\end{tabular}
\begin{tablenotes}
    \footnotesize
    \item * The performance of each category refers to the average detection accuracy of all model-specific categories under it.
    \item * GFLOPs are computed by the built-in analysis tool of MMEngine. 
    \item * ``R.'' denotes the rotated version of the detector. 
\end{tablenotes}
\end{threeparttable}
\vspace{-2mm}
\end{table*}

\subsection{Compare with State-of-the-Art Methods}
We compare the proposed ExpertDet with a wide range of object detectors on PSP.Ship and PSP.Plane, including general-purpose, fine-grained, and open-vocabulary methods, spanning one-stage, two-stage, and transformer-based architectures.
In addition, we compare ExpertDet with state-of-the-art (SOTA) fine-grained detectors on several existing fine-grained datasets, i.e., MAR20, ShipRSImageNet, and FAIR1M-v2, to further evaluate its effectiveness and generalization capability.

\begin{figure*}
  \centering
  \includegraphics[width=0.95\linewidth]{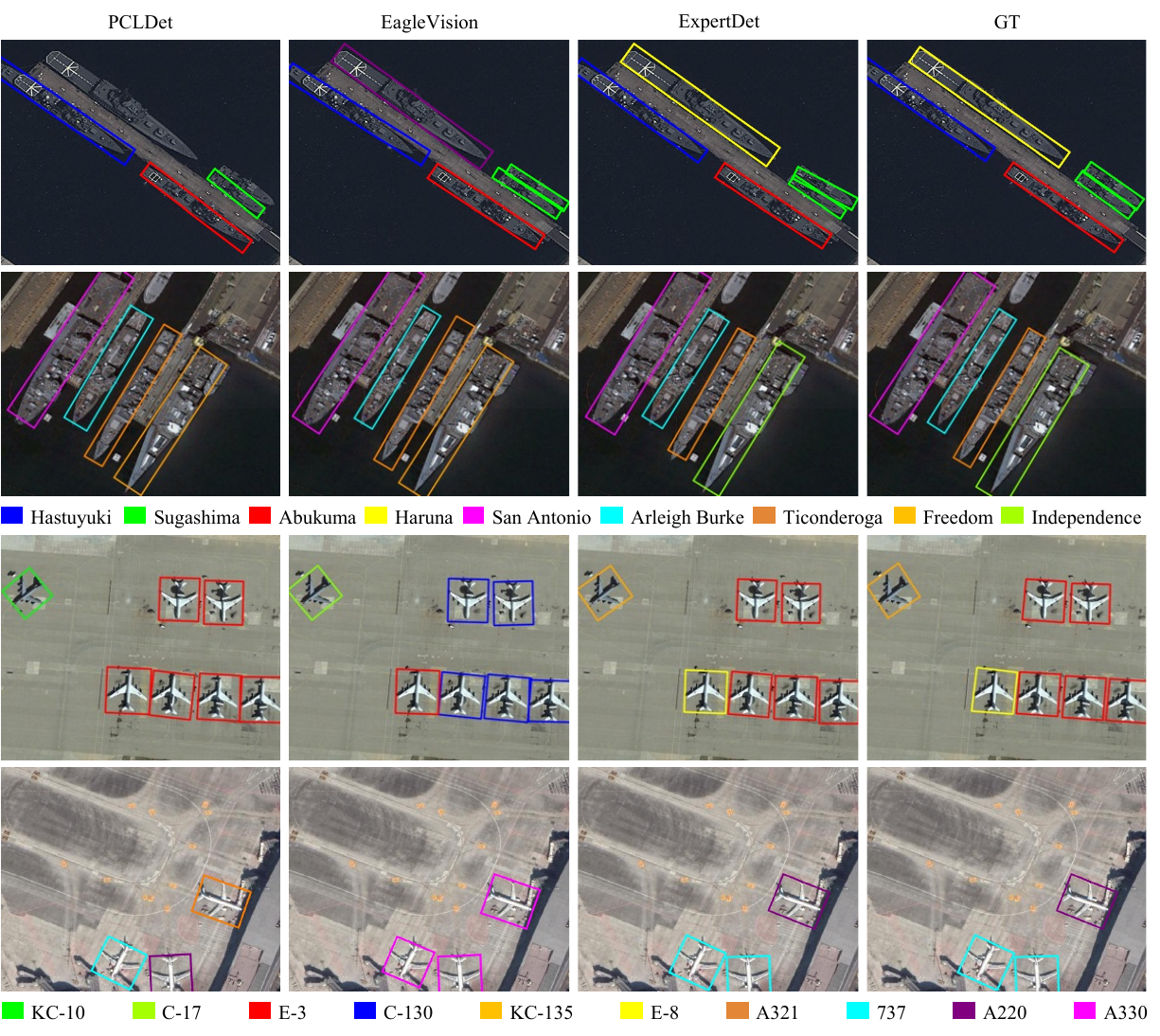}
  \caption{Visualization results of PCLDet, EagleVision, and ExpertDet. The first two rows are from PSP.Ship, and the last two rows are from PSP.Plane.}
  \label{fig:det}
\vspace{-4mm}
\end{figure*}

\textbf{Results on PSP.Ship}.
The results are reported in Table~\ref{tab:overall_ship}. 
ExpertDet achieves consistent performance gains over the existing single-stage detector R. ATSS\cite{zhang2020atss} as well as the two-stage detectors ReDet~\cite{han2021redet} and Oriented R-CNN~\cite{xie2021oriented}. Among them, ExpertDet built upon Oriented R-CNN achieves the best overall performance.
Detailed analysis is presented below.

\textit{Different Detection Paradigms}.
We first evaluate the overall performance of different detection paradigms on PSP.Ship, including one-stage, Transformer-based and two-stage detectors. 
The two-stage detectors generally exhibit superior performance compared to one-stage and Transformer-based detectors.
In particular, the top-performing two-stage method, Oriented R-CNN~\cite{xie2021oriented}, surpasses the best one-stage method, R. ATSS~\cite{zhang2020atss}, by approximately 18 points, and the best Transformer-based method, RHINO~\cite{lee2024hausdorff}, by around 15 points.
It can be primarily attributed to the fact that PSP.Ship contains numerous model-specific categories with subtle inter-model variations, requiring detectors to rely on higher-quality instance features to learn robust fine-grained representations.
Two-stage methods leverage the Region Proposal Network (RPN) to explicitly sample Regions of Interest (RoIs) and utilize Intersection over Union (IoU) thresholds to provide the second-stage classifier with high-quality instance features, thereby effectively capturing subtle differences between object models.
In contrast, one-stage methods perform classification and regression directly on the entire feature map. The long-tail distribution and sample scarcity further exacerbate the inherent foreground-background imbalance in such frameworks, making it significantly more challenging to learn stable fine-grained decision boundaries. 
Transformer-based detectors, on the other hand, rely on Hungarian matching to enforce one-to-one assignment, resulting in inherently sparse positive supervision, and thus tend to be less effective on PSP.Ship, where most ship models contain only a limited number of samples.

Notably, GLIP~\cite{li2022glip} and Grounding DINO~\cite{liu2024grounding} are two representative open-vocabulary detectors that leverage large-scale vision–language pretraining to enable semantic generalization under limited data. Nevertheless, their performance remains suboptimal on the fine-grained PSP.Ship benchmark, as the text descriptions used during pretraining typically focus on category-level semantics, providing insufficient discriminative cues for distinguishing subtle inter-model differences.

\textit{Long-Tailed Object Detection Methods}.
We then evaluate two detectors specifically designed to address the long-tailed distribution problem, i.e., EQLv2~\cite{tan2023eql} and LogN~\cite{zhao2024logit}.
However, these methods yield suboptimal performance when the long-tailed distribution is coupled with fine-grained detection and sample scarcity.
EQLv2 alleviates the training bias caused by long-tailed distributions by suppressing negative gradients on tail categories. 
While this strategy is generally effective on conventional long-tailed benchmarks with relatively large inter-category discrepancies, it is less suitable for PSP.Ship, where categories are visually highly similar.
Many semantically adjacent categories actually serve as hard negatives, which are essential for learning subtle decision boundaries.
Suppressing such negative gradients weakens important discriminative signals, thereby making the distinction among fine-grained categories more difficult.
LogN assumes that long-tail bias can be characterized by class-wise logit statistics and corrected through logit normalization. 
Nevertheless, since a vast majority of tail categories in PSP.Ship contain only a very limited number of training instances, making the estimated logit statistics less stable and fail to accurately represent the underlying semantic distribution of categories.
Consequently, LogN struggles to effectively correct the long-tail bias, and may even introduce inaccurate calibration that further distorts the decision boundaries between visually similar categories, leading to degraded performance.

\textit{Fine-Grained Object Detection Methods}.
We further evaluate a series of representative fine-grained object detection methods, including OM~\cite{zhu2024enhancing}, ISCL~\cite{zeng2022instance}, HiCLPL~\cite{zhang2022hierarchical}, PETDet~\cite{li2023petdet}, PCLDet~\cite{ouyang2023pcldet}, and EagleVision~\cite{jiang2025eaglevision}. 
Experimental results show that these methods generally underperform their corresponding baseline detectors on PSP.Ship.
It suggests that in fine-grained scenarios characterized by a massive scale of model-specific categories with highly dense decision boundaries, the assumptions underlying existing fine-grained methods become less effective.
For example, PETDet utilizes multi-scale feature learning to capture fine-grained structural differences. However, on PSP.Ship, the differences between ship models are often extremely subtle and highly localized. Although multi-scale aggregation introduces richer contextual information, it may also dilute critical local discriminative cues, thereby degrading performance.

OM, ISCL, and PCLDet mainly improve representation separability by enlarging inter-category margins. 
However, on PSP.Ship, the number of model-specific categories is much larger and the category space is highly crowded. 
Under such conditions, indiscriminately forcing all categories apart not only exceeds the effective representational capacity of the feature space, but also disrupts the originally reasonable semantic relationships among hierarchically related categories. 
As a result, these methods struggle to establish stable fine-grained decision boundaries, leading to degraded performance. 
Notably, OM yields a marginal gain in one-stage settings. It is considered that the dense prediction head of a one-stage architecture has to simultaneously handle severe foreground-background imbalance and category coupling across the entire feature map. In such a chaotic feature space, the strict orthogonal constraint serves as a robust separation prior, partially reducing classification confusion. By contrast, in two-stage settings, the classification head operates on refined RoI features with higher semantic purity, where such rigid constraints tend to act as over-regularization, thereby hindering the modeling of subtle yet meaningful variations among related models.

HiCLPL explicitly introduces hierarchical relations to enhance separability. 
However, it suffers from a substantial performance drop relative to its base detector on PSP.Ship.
It is considered that HiCLPL transforms the flat classification layer into a binary classification tree, which tends to introduce noticeable cascading errors, \ie, an incorrect prediction at an early higher hierarchy level may directly block the subsequent correct fine-grained decision path.
In addition, HiCLPL enforces the alignment of bottom-level specific models with higher-level semantic clusters, leading to feature homogenization.
As the tree grows in depth and breadth, these limitations become increasingly pronounced, complicating optimization and leading to suboptimal convergence.

EagleVision is an advanced MLLM that unifies object detection and attribute comprehension within a single framework to promote mutual enhancement, whose effectiveness inherently relies on massive amounts of high-quality image-text paired data.
Under conditions of sample scarcity, EagleVision lacks sufficient instances to establish accurate vision-language alignment. As a result, instead of providing beneficial semantic priors, it may introduce substantial semantic noise that interferes with the base detection task and degrades its ability to distinguish subtle inter-model visual differences.

\textit{Superiority of ExpertDet}.
ExpertDet introduces attribute supervision through masked attribute reconstruction, which imposes a more constrained task space and a clearer supervisory objective, thereby enabling more stable optimization. By requiring the recovery of only masked attributes, it directs the model more explicitly toward the fine-grained discriminative cues that determine inter-model differences. Moreover, instead of exploiting hierarchical relations through a top-down conditional decision process, ExpertDet incorporates them via a visual prototype tree with hierarchical contrastive learning, thereby preserving semantic continuity across levels while preventing the homogenization of fine-grained features. 
Consequently, ExpertDet is more compatible with the highly crowded category space of PSP.Ship and attains superior performance. 
We visualize two representative ship detection results in Fig.~\ref{fig:det}, which qualitatively demonstrate the superiority of ExpertDet over state-of-the-art fine-grained detectors in terms of discriminating ship models.

\textbf{Results on PSP.Plane}.
The results are reported in Table.~\ref{tab:overall_plane}.
Similar to the observations on PSP.Plane, ExpertDet consistently outperforms all baseline detectors and specialized fine-grained methods. 
Compared with PSP.Ship, however, the performance gap between two-stage, one-stage and Transformer-based detectors narrows significantly, as PSP.Plane features fewer model-specific categories and more abundant samples per class, effectively mitigating the foreground-background imbalance inherent in one-stage frameworks.
Meanwhile, the performance degradation of both long-tailed and fine-grained detection methods relative to their baseline detectors is notably mitigated. 
ExpertDet consistently achieves performance improvements across both datasets, indicating its robustness to variations in category scale and sample distribution.

\begin{table*}[!bt]
\setlength{\tabcolsep}{4.5pt}
\centering
\caption{Experimental results of general-purpose, fine-grained, and open-vocabulary object detection methods on PSP.Plane, covering one-stage, two-stage, and Transformer-based architectures.}
\label{tab:overall_plane}
\begin{tabular}{l|c|cc|cccccccccccc}
\toprule
Method & Airplane & M         & C       & F    & B    & FB   & MC   & T    & EWC  & AS   & GSBM & RT   & NBA  & WBA  & RA   \\
\midrule
\rowcolor{gray!20}
\multicolumn{1}{l}{\textit{One-stage}} & \multicolumn{1}{c}{}&\multicolumn{1}{c}{} & \multicolumn{1}{c}{}& \multicolumn{1}{c}{}& & & & & & \multicolumn{1}{c}{}& & &&& \\
R. RetinaNet~\cite{lin2020retinanet} & 58.5      & 72.5               & 30.6          & 64.4 & 82.8 & 74.6 & 76.9 & 69.7 & 82.4 & 77.1 & 58.3 & 58.8 & 27.7 & 35.0 & \underline{20.4} \\
R. FCOS~\cite{tian2019fcos} & 61.9      & 76.9               & 31.8          & 64.2 & 86.0 & 74.1 & 80.9 & 78.0 & 82.2 & 88.4 & 69.8 & 68.1 & 28.3 & 37.3 & 18.0 \\
R. ATSS~\cite{zhang2020atss} & 62.6      & 77.6               & 32.8          & 65.1 & 85.6 & 76.4 & 79.7 & 77.6 & 85.0 & 84.2 & 67.4 & 76.2 & 32.9 & 35.9 & 16.6 \\
R$^3$Det~\cite{yang2021r3det} & 56.5      & 71.6               & 26.4          & 68.7 & 81.2 & 75.1 & 83.2 & 71.3 & 59.9 & 71.8 & 49.2 & 58.0 & 25.3 & 30.3 & 11.3 \\
S$^2$A-Net~\cite{han2021align} & 62.6      & 79.6               & 28.7          & 70.3 & 84.2 & 78.0 & 85.3 & 78.6 & 85.1 & 84.0 & 77.4 & 79.7 & 24.1 & 35.4 & 14.2 \\
R. GLIP~\cite{li2022glip} & 58.4      & 72.4               & 30.3          & 64.2 & 82.9 & 78.4 & 80.2 & 72.5 & 68.9 & 84.4 & 71.4 & 46.9 & 28.6 & 34.4 & 16.9 \\
OM (R. FCOS)~\cite{zhu2024enhancing} & 64.2 \scriptsize{\mygreen{(+2.3)}} & 80.5               & 31.7          & 72.1 & 85.5 & 77.9 & 85.9 & 83.6 & 88.3 & 89.0 & 66.8 & 75.4 & 32.4 & 33.5 & 19.9 \\
\midrule
\rowcolor{gray!20}
\multicolumn{1}{l}{\textit{Transformer-based}} & \multicolumn{1}{c}{}&\multicolumn{1}{c}{} & \multicolumn{1}{c}{}& \multicolumn{1}{c}{}& & & & & & \multicolumn{1}{c}{}& &&&&\\
R. DeformableDETR~\cite{zhu2021deformable} & 51.8      & 65.0               & 25.6          & 56.8 & 71.0 & 65.4 & 71.9 & 69.3 & 67.1 & 71.0 & 59.3 & 51.6 & 20.7 & 32.3 & 11.8 \\
RHINO~\cite{lee2024hausdorff} & 60.0      & 76.2               & 27.8          & 63.5 & 82.9 & 75.7 & 79.1 & 78.3 & 82.1 & 80.2 & 73.9 & 73.3 & 25.8 & 31.8 & 15.6 \\
R. Grounding DINO~\cite{liu2024grounding} & 59.3      & 74.7               & 28.3          & 66.8 & 82.5 & 75.4 & 85.2 & 74.5 & 78.0 & 76.6 & 56.0 & 68.8 & 26.6 & 31.8 & 17.8 \\
\midrule
\rowcolor{gray!20}
\multicolumn{1}{l}{\textit{Two-stage}} & \multicolumn{1}{c}{}&\multicolumn{1}{c}{} & \multicolumn{1}{c}{}& \multicolumn{1}{c}{}& & & & & & \multicolumn{1}{c}{}& &&&&\\
R. Faster R-CNN~\cite{ren2016faster} & 61.1      & 78.6               & 26.3          & 67.1 & 82.1 & 72.9 & 86.7 & 83.6 & 87.8 & 88.7 & 75.3 & 74.0 & 23.2 & 31.0 & 15.0 \\
Oriented R-CNN~\cite{xie2021oriented} & 66.4      & 84.7               & 29.8          & 74.7 & \underline{88.8} & 82.9 & 88.4 & 85.8 & 90.1 & 90.2 & 81.7 & 86.5 & 29.2 & 33.3 & 14.8 \\
ReDet~\cite{han2021redet} & \underline{67.9}      & 85.7               & 32.4          & 75.1 & \textbf{88.9} & \underline{86.3} & 88.3 & 86.4 & 90.1 & \underline{90.8} & \textbf{83.0} & \textbf{91.3} & 30.7 & 37.0 & 15.6 \\
Gliding Vertex~\cite{xu2021gliding} & 57.7 & 73.9 & 25.4 & 64.7 & 78.2 & 69.8 & 78.7 & 78.6 & 84.0 & 77.7 & 66.5 & 68.4 & 22.4 & 29.3 & 17.7 \\
EQLv2~\cite{tan2023eql} (Oriented R-CNN) & 66.1 \scriptsize{\myred{(-0.3)}} & 85.1 & 28.2 & 75.2 & 88.3 & 82.9 & 88.5 & 88.7 & \underline{90.4} & 89.2 & 80.9 & 86.4 & 26.0 & 32.4 & 16.3 \\
LogN~\cite{zhao2024logit} (Oriented R-CNN) & 64.4 \scriptsize{\myred{(-2.0)}} & 83.5 & 26.3 & 74.2 & 88.4 & 78.3 & \underline{89.4} & 85.6 & 89.5 & 90.2 & 77.6 & 80.3 & 26.0 & 29.2 & 13.4 \\
R. HiCLPL~\cite{zhang2022hierarchical} (R. Faster R-CNN) & 59.2 \scriptsize{\myred{(-1.9)}} & 75.3 & 26.9 & 63.0 & 77.4 & 66.3 & 78.3 & 82.0 & 85.1 & 77.7 & 77.1 & 79.2 & 23.9 & 30.9 & 19.3  \\
ISCL~\cite{zeng2022instance} (Oriented R-CNN) & 65.1 \scriptsize{\myred{(-1.4)}} & 82.2               & 30.8          & 72.5 & 86.2 & 79.6 & 85.3 & 86.7 & \textbf{90.5} & 89.3 & 68.4 & 82.0 & 27.2 & 35.4 & \textbf{21.9} \\
PETDet~\cite{li2023petdet} (Oriented R-CNN) & 63.1 \scriptsize{\myred{(-3.3)}} & 79.5 & 30.4 & 69.8 & 85.9 & 77.1 & 86.0 & 80.2 & 89.0 & 86.6 & 72.6 & 71.6 & 30.4 & 33.2 & 16.7 \\
PCLDet~\cite{ouyang2023pcldet} (Oriented R-CNN) & 66.3 \scriptsize{\myred{(-0.1)}} & 84.4 & 30.2 & 73.2 & 88.5 & 81.3 & 81.4 & \textbf{88.7} & 89.6 & 89.7 & 82.0 & 86.8 & 27.9 & 35.1 & 15.2 \\
OM~\cite{zhu2024enhancing} (Oriented R-CNN) & 66.3 \scriptsize{\myred{(-0.1)}} & 84.3 & 30.2 & 75.5 & 88.5 & 80.1 & 88.7 & 88.6 & 90.5 & 89.2 & 74.2 & 82.6 & 30.2 & 33.8 & 12.1 \\
EagleVision~\cite{jiang2025eaglevision} (Oriented R-CNN)  & 66.3 \scriptsize{\myred{(-0.1)}} & 84.1 & 30.6 & 71.9 & 88.2 & 81.2 & 88.7 & 86.3 & 89.2 & 90.3 & 80.1 & 88.7 & 28.3 & 34.0 & 22.9 \\
\midrule
\rowcolor{gray!20}
\multicolumn{1}{l}{\textit{Ours}} & \multicolumn{1}{c}{}&\multicolumn{1}{c}{} & \multicolumn{1}{c}{}& \multicolumn{1}{c}{}& & & & & & \multicolumn{1}{c}{}& &&&&\\
\rowcolor{blue!5}
ExpertDet (R. ATSS)      & 64.9 \scriptsize{\mygreen{(+2.3)}} & 80.5 & \underline{33.8} & 69.3 & 86.1 & 77.1 & 84.0 & 82.6 & 85.2 & 88.3 & 82.0 & 75.9 & \underline{31.2} & \textbf{39.5} & 16.2 \\
\rowcolor{blue!5}
ExpertDet (ReDet)          & \textbf{68.8} \scriptsize{\mygreen{(+0.9)}} & \textbf{86.1} & \textbf{34.1} & \underline{77.3} & \textbf{88.9} & \textbf{89.0} & \textbf{89.2} & 86.7 & 89.9 & \textbf{90.9} & \underline{82.3} & \underline{89.1} & \textbf{31.6} & \underline{39.2} & 18.6 \\
\rowcolor{blue!5}
ExpertDet (Oriented R-CNN) & 67.6 \scriptsize{\mygreen{(+1.2)}}& \underline{86.0} & 30.9 & \textbf{77.8} & \textbf{88.6} & 83.4 & 88.2 & \underline{88.9} & 90.5 & 90.3 & 83.2 & 87.2 & 28.0 & 35.6 & 19.0 \\
\bottomrule
\end{tabular}
\vspace{-2mm}
\end{table*}

\textbf{Results on existing fine-grained datasets}.
To further evaluate the generalization capability of the proposed method, we compare ExpertDet with SOTA methods on three existing fine-grained datasets, MAR20, ShipRSImageNet, and FAIR1M-v2.
ShipRSImageNet and FAIR1M-v2 mainly consist of subcategories, with only a subset corresponding to specific models. 
Specifically, ShipRSImageNet contains 50 categories, among which 21 are model-specific categories, while FAIR1M-v2 contains 37 categories, among which 10 are model-specific categories.
During training, ExpertDet applies available expert knowledge guided supervision to model-specific categories.
The results are shown in Tab.~\ref{tab:overall_others}.
ExpertDet consistently achieves the best performance across all three datasets, demonstrating its strong generalization ability on existing fine-grained benchmarks.
And we can observe that existing methods show competitive effectiveness on benchmarks such as ShipRSImageNet and FAIR1M-v2, where the number of categories is relatively limited and the category granularity is coarser.
However, their advantages become less consistent on large-scale model-specific benchmark PSP, where the category space is denser and the inter-model differences are more subtle.
It further highlights the significance of PSP, which provides more challenging benchmarks for evaluating fine-grained detection capability and better aligns with the requirements of real-world precision applications.

\begin{table}[!bt]
    \centering
    \caption{Comparison of SOTA fine-grained detectors and ExpertDet on existing fine-grained datasets, including MAR20, ShipRSImageNet and FAIR1M-v2.}
    \label{tab:overall_others}
    \begin{tabular}{lccc} 
        \toprule
        Methods & MAR20 & ShipRSImageNet & FAIR1M-v2 \\
        \midrule
        Oriented R-CNN~\cite{xie2021oriented} & 85.8 & 73.8 & 41.2 \\
        PETDet~\cite{li2023petdet} & 84.1 & 75.6 & 44.3 \\
        PCLDet~\cite{ouyang2023pcldet} & 85.6 & 74.4 & 39.8 \\
        OM~\cite{zhu2024enhancing} & 84.7 & 75.0 & \textbf{44.5} \\
        EagleVision~\cite{jiang2025eaglevision} & 84.7 & 72.5 & 40.8 \\
        \rowcolor{gray!20}
        \textbf{ExpertDet} & \textbf{86.2} & \textbf{76.2} & \textbf{44.5} \\ 
        \bottomrule
    \end{tabular}
\vspace{-2mm}
\end{table}

\subsection{Ablation Studies}
\textbf{Individual Effectiveness of VMAM and HierVIP}.
We evaluate the individual effectiveness of VMAM and HierVIP on the PSP.Ship and PSP.Plane datasets, as shown in Tab.~\ref{tab:ablation}.
Either VMAM or HierVIP alone yields consistent performance improvements, indicating that both attributes and hierarchies provide effective supervision for fine-grained detection.
Combining the two complementary forms of expert knowledge leads to more substantial gains.
Moreover, the improvement on PSP.Ship is more pronounced than that on PSP.Plane, since PSP.Ship contains more categories with fewer samples per category, where category-label supervision alone is insufficient for learning stable fine-grained decision boundaries, and expert knowledge provides additional constraints that enhance fine-grained separability.

\begin{table}[!bt]
\setlength{\tabcolsep}{4pt}
  \caption{Individual Effectiveness of VMAM and HierVIP on PSP.Ship and PSP.Plane.}
  \label{tab:ablation}
  \centering
  \begin{tabular}{c|cc|cc}
    \toprule
    Methods & VMAM & HierVIP & PSP.Ship & PSP.Plane \\
    \midrule
    Baseline & & & 79.3 & 66.4  \\
    \midrule 
    \multirow{3}{*}{ExpertDet} & \ding{51} & & 81.2 & 66.7 \\
    & & \ding{51} & 80.9 & 66.8 \\ 
    & \ding{51} & \ding{51} & \textbf{82.5} & \textbf{67.6} \\
  \bottomrule
  \end{tabular}
\vspace{-2mm}
\end{table}

\begin{figure}[!t]
  \centering
  \includegraphics[width=0.99\linewidth]{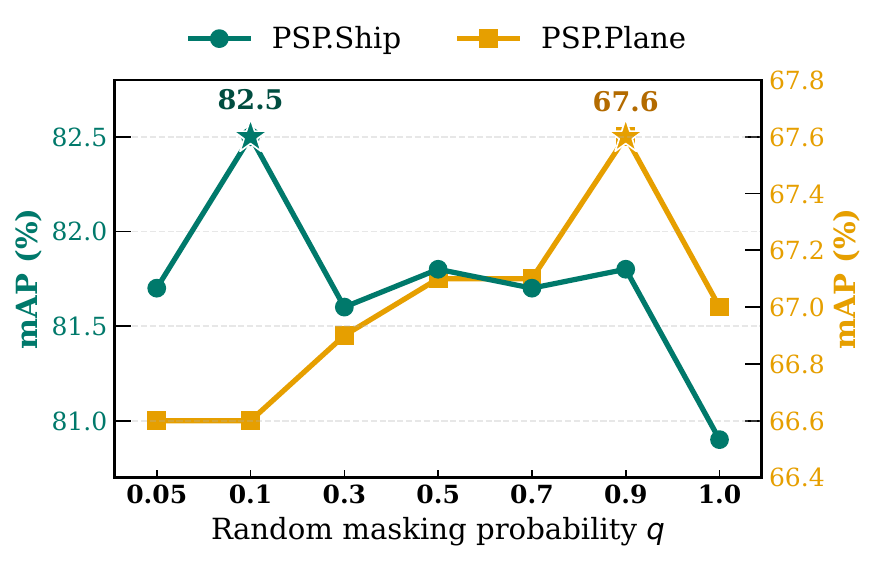}
  \caption{Effect of random masking probability in VMAM on PSP.Ship and PSP.Plane.}
  \label{fig:vmam_q}
\vspace{-4mm}
\end{figure}

\noindent\textbf{Effect of Random Masking Probability in VMAM.}
VMAM aligns attribute semantics with visual structures by reconstructing randomly masked attributes from visual cues, enabling the detector to capture subtle structural distinctions.
We study the effect of different masking probabilities on PSP.Ship and PSP.Plane.
The experimental results are illustrated in Fig.~\ref{fig:vmam_q}, where the optimal performance is achieved at $q=0.1$ for PSP.Ship and $q=0.9$ for PSP.Plane.
Given the inherent sample scarcity in the PSP.Ship dataset, a smaller masking probability provides a more stable reconstruction objective.
As $q$ increases, the reconstruction task becomes excessively challenging due to the loss of semantic anchors. 
In contrast, PSP.Plane contains relatively more samples, where a higher masking probability can strengthen the grounding of attribute semantics.
When $q=1$, VMAM degenerates into an open-ended generation task that attempts to recover the entire attribute description from visual features, making the predicted attributes less reliably grounded in structured semantics and weakening the precise alignment between attribute semantics and visual structures, thereby degrading performance.
\begin{table}[!t]
  \caption{Effect of different attributes in VMAM on PSP.Ship.}
  \label{tab:attr}
  \centering
  \setlength{\tabcolsep}{4.5pt}
  \begin{tabular}{c|c|ccccccc}
    \toprule
    Attributes & Ship    & SC   & AW   & CG   & AS   & MW   & AC   \\
    \midrule
    None        & 80.9 & 81.3 & 68.5 & 67.9 & 60.5 & 55.4 & 93.7 \\ 
    Absolute Size Cues   & 81.5 & 81.3 & 67.5 & 70.0 & 61.5 & 63.2 & 90.5 \\
    Stylistic Semantic Cues    & 81.3 & 79.5 & 68.3 & 73.6 & 60.8 & 58.2 & 93.9 \\
    Structural Detail Cues   & 81.8 & 81.4 & 68.4 & 69.7 & 62.0 & 61.5 & 90.1 \\
    Random & 81.8 & 81.4 & 67.9 & 69.2 & 61.5 & 61.9 & 93.7 \\
    All & 82.5 & 81.1 & 68.5 & 68.5 & 63.4 & 60.3 & 93.9 \\
  \bottomrule
  \end{tabular}
\vspace{-2mm}
\end{table}

\noindent\textbf{Effect of Different Attributes in VMAM.}
We also compare the performance of ExpertDet using different attributes on the PSP.Ship dataset.
For each model category in PSP.Ship, the attribute entries are organized from three perspectives: absolute size cues (\textit{overall length} and \textit{beam}), stylistic semantic cues (\textit{operator}), and structural detail cues (\textit{bow}, \textit{stern}, \textit{midships structure}, \textit{funnel}, \textit{fore bulwark}, \textit{vertical replenishment point}, \textit{gun type}, and \textit{missile launcher type}).
The results are reported in Tab.~\ref{tab:attr}.
We can observe that all types of attribute cues bring performance improvements, indicating that different attributes provide effective semantic supervision for fine-grained detection.
Among them, structural detail cues contribute the most significant gains, since model-specific ship categories are primarily distinguished by subtle local structures and equipment configurations.
Furthermore, we evaluate a \textit{random} setting where only 8 out of 11 attributes are stochastically selected for each model category during training, which simulates realistic scenarios where expert knowledge may be incomplete or partially inaccessible. We observe that even with restricted attribute availability, ExpertDet still maintains superior performance compared to the baseline,  indicating that it can effectively exploit fragmented semantic cues to enhance fine-grained discrimination.
\begin{table}[!t]
  \caption{Effect of the HSC Loss on PSP.Ship.}
  \label{tab:hier_csl}
  \centering
  \begin{tabular}{c|c|ccccccc}
    \toprule
    Loss & Ship    & SC   & AW   & CG   & AS   & MW   & AC   \\
    \midrule
    None & 81.2 & 80.9 & \textbf{69.1} & 70.3 & 60.9 & 52.8 & 93.8 \\
    SupCon~\cite{khosla2020supervised} & 80.6 & 81.0 & 68.3 & \textbf{72.1} & 58.9 & 64.1 & 90.0 \\
    HiCL~\cite{zhang2022hierarchical}  & 81.4 & \textbf{81.1} & 67.1 & 71.3 & 61.3 & \textbf{65.8} & 88.9 \\
    HSC (Ours) & \textbf{82.5} & \textbf{81.1} & 68.5 & 68.5 & \textbf{63.4} & 60.3 & \textbf{93.9} \\
  \bottomrule
  \end{tabular}
\vspace{-2mm}
\end{table}

\begin{figure}[!t]
  \centering
  \includegraphics[width=0.99\linewidth]{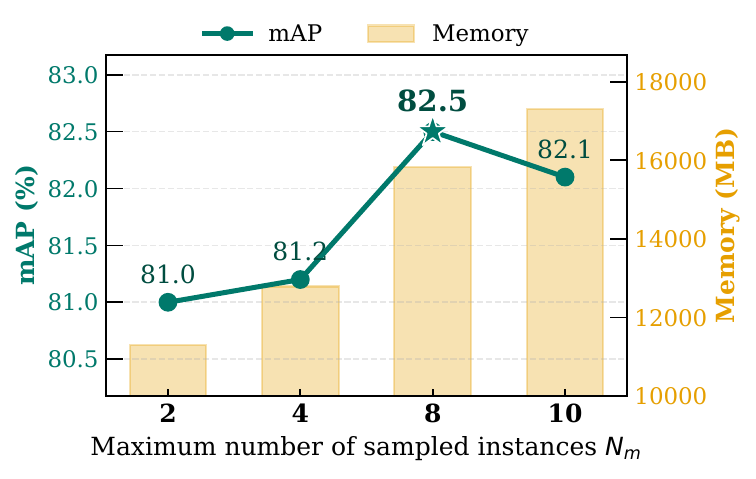}
  \caption{Effect of maximum instance number in VMAM on PSP.Ship.}
  \label{fig:vmam_m}
\vspace{-4mm}
\end{figure}
\noindent\textbf{Effect of Maximum Instance Number in VMAM.}
We further investigate the influence of $N_m$, which specifies the number of instances sampled for masked attribute reconstruction in VMAM, on the PSP.Ship dataset.
The results are shown in Fig.~\ref{fig:vmam_m}.
Increasing $N_m$ generally improves performance by providing denser supervision for aligning attribute semantics with visual structures, at the cost of higher memory consumption, with the best performance obtained at $N_m=8$.

\noindent\textbf{Effect of the HSC Loss. }
The HSC loss is introduced in HierVIP to optimize the visual prototype tree, aiming to improve feature separability under the hierarchical structure while preserving semantic continuity across levels.
To evaluate its effectiveness, we compare the HSC loss with two alternative contrastive objectives, \ie{}, SupCon~\cite{khosla2020supervised} and HiCL~\cite{zhang2022hierarchical}, as the supervision objective over the prototype tree.
Specifically, SupCon does not explicitly model hierarchical relations and treats all model-specific categories as flat and mutually exclusive classes, while HiCL introduces hierarchy-aware supervision but performs contrastive learning across both intra-level and inter-level categories.
The results are reported in Tab.~\ref{tab:hier_csl}.
We observe that SupCon leads to a certain performance drop, suggesting that, in scenarios with numerous and highly similar model-specific categories, uniformly separating semantically related models in a flat feature space may cause feature collapse.
HiCL brings moderate improvements, indicating the usefulness of hierarchical information.
However, its effectiveness is limited by the conflict introduced by directly contrasting categories across different semantic granularities.
In contrast, the proposed HSC loss preserves the parent-child relations in the prototype tree while restricting contrastive learning to categories within the same hierarchy level, resulting in a more stable and discriminative feature space.

\noindent\textbf{Effect of Adaptive Momentum in HierVIP.}
We further evaluate the effectiveness of the adaptive momentum strategy used for updating the visual prototype tree in HierVIP on the PSP.Ship dataset, with results reported in Tab.~\ref{tab:adp_csl}.
Compared with static momentum, adaptive momentum yields better performance by adjusting prototype updates according to feature drift, thereby improving both adaptability and stability.

\begin{table}[!t]
  \caption{Effect of adaptive momentum in HierVIP on PSP.Ship.}
  \label{tab:adp_csl}
  \centering
  \setlength{\tabcolsep}{4.5pt}
  \begin{tabular}{c|c|ccccccc}
    \toprule
    Strategy & Ship    & SC   & AW   & CG   & AS   & MW   & AC   \\
    \midrule
    Static Momentum  & 81.9 & \textbf{81.4} & 67.5 & \textbf{68.9} & 62.3 & 59.9 & 93.5 \\
    Adaptive Momentum & \textbf{82.5} & 81.1 & \textbf{68.5} & 68.5 & \textbf{63.4} & \textbf{60.3} & \textbf{93.9} \\
  \bottomrule
  \end{tabular}
\vspace{-2mm}
\end{table}

\noindent\textbf{Effect of Weighting Coefficients. } 
We investigate the effect of the weighting coefficients in the loss function, \ie{}, $\lambda$ and $\mu$, as shown in Tab.~\ref{tab:w}. 
We can find that $\lambda$ and $\mu$ remain robust within a certain range, with the best performance achieved
when $\lambda=0.20$ and $\mu=0.001$.


\begin{table}[!t]
    \centering
    \caption{Effect of weighting coefficients on PSP.Ship.}
    \label{tab:w}
    \setlength{\tabcolsep}{4pt}
    \begin{tabular}{c|cccc|cccc}
        \toprule
        $\lambda$   & \multicolumn{4}{c|}{0.1} & \multicolumn{4}{c}{0.15} \\ \midrule
        $\mu$   & 0.0005 & 0.001 & 0.0025 & 0.005 & 0.0005 & 0.001 & 0.0025 & 0.005 \\ \midrule
        mAP   & 80.8 & 82.1 & 80.6 & 80.0 & 80.6 & 80.4 & 80.2 & 82.4 \\ 
        \midrule \midrule
        $\lambda$   & \multicolumn{4}{c|}{0.2} & \multicolumn{4}{c}{0.25} \\ \midrule
        $\mu$   & 0.0005 & 0.001 & 0.0025 & 0.005 & 0.0005 & 0.001 & 0.0025 & 0.005 \\ \midrule
        mAP   & 81.9 & \textbf{82.5} & 80.7 & 81.9 & 80.2 & 81.0 & 81.7 & 81.7 \\ 
        \bottomrule
    \end{tabular}
\vspace{-2mm}
\end{table}


\subsection{Analysis}

\noindent\textbf{Performance on sample-scarce categories. }
In real-world scenarios, fine-grained object detection often suffers from long-tailed category distributions, uneven category occurrence frequencies, and costly expert annotation, making sufficient and reliable training samples difficult to obtain for many categories.
We therefore evaluate the performance on 17 tail categories in the PSP.Ship dataset, each containing fewer than 10 training instances, as shown in Tab.~\ref{tab:small_sample}.
The results demonstrate that ExpertDet achieves the best overall performance, with largely consistent gains across the long-tailed category distribution, suggesting that expert-knowledge guided supervision provides effective auxiliary constraints for sample-scarce categories and enables more reliable fine-grained discrimination under limited training data.


\begin{table}[!t]
  \centering
  \caption{Comparison of ExpertDet with SOTA methods on sample-scarce categories in PSP.ship.}
\setlength{\tabcolsep}{2.8pt}
  \label{tab:small_sample}
    \begin{tabular}{lcccc}
      \toprule
      Category & Oriented R-CNN & PCLDet & EagleVision & ExpertDet \\
      \midrule
      Anchorage & 31.8 & \textbf{33.3} & 30.3 & 27.3 \\
      Blue Ridge & \textbf{100.0} & \textbf{100.0} & 57.6 & \textbf{100.0} \\
      Chiyoda & \textbf{100.0} & \textbf{100.0} & 0.0 & \textbf{100.0} \\
      Endurance & \textbf{100.0} & \textbf{100.0} & \textbf{100.0} & \textbf{100.0} \\
      Enterprise & 36.4 & 27.3 & 23.0 & \textbf{54.5} \\
      Futami & \textbf{13.6} & 9.1 & 0.0 & \textbf{13.6} \\
      Guam & \textbf{100.0} & \textbf{100.0} & 13.6 & \textbf{100.0} \\
      John F Kennedy & \textbf{100.0} & \textbf{100.0} & \textbf{100.0} & \textbf{100.0} \\
      Kuznetsov & \textbf{100.0} & \textbf{100.0} & 54.5 & \textbf{100.0} \\
      Newport & 87.4 & 69.1 & 85.5 & \textbf{100.0} \\
      Nichinan & 0.0 & \textbf{21.4} & 9.1 & 18.2 \\
      Pigeon & 0.0 & 0.0 & 0.0 & \textbf{100.0} \\
      Príncipe de Asturias & \textbf{100.0} & \textbf{100.0} & 50.0 & \textbf{100.0} \\
      Sacramento & \textbf{54.5} & \textbf{54.5} & \textbf{54.5} & \textbf{54.5} \\
      Shirase & \textbf{100.0} & \textbf{100.0} & \textbf{100.0} & \textbf{100.0} \\
      Type C8 & \textbf{100.0} & \textbf{100.0} & 79.2 & \textbf{100.0} \\
      Victorious & 0.0 & 25.0 & \textbf{100.0} & \textbf{100.0} \\
      \midrule
      Mean & 72.0 & 65.6 & 50.4 & \textbf{80.5} \\
      \bottomrule
    \end{tabular}
\end{table}

\noindent\textbf{Performance at Coarser Hierarchy Levels.}
In practical applications, different tasks may require different levels of recognition granularity. 
Beyond model-level evaluation, we further evaluate detection performance at coarser hierarchy levels on PSP.Ship and PSP.Plane.
Specifically, a prediction is considered correct if it matches the ground-truth label after both are mapped to the target hierarchy level, \ie{}, confusions among model-specific categories belonging to the same target-level category are ignored.
We denote the mAP at hierarchy level $l$ as $\mathrm{mAP}^{l}$, where $\mathrm{mAP}^{0}$ corresponds to model-level evaluation and larger $l$ indicates progressively coarser levels.
The results are reported in Tab.~\ref{tab:coarse_results}.
We observe that existing fine-grained methods, such as PCLDet and OM, underperform their corresponding baseline detectors at the model level, yet yield marginal improvements at coarser hierarchical levels.
It suggests that the underlying assumptions of these methods, while potentially effective for for scenarios with fewer categories and more distinguishable inter-class variations, fail to scale to a massive category space characterized by highly dense decision boundaries.
In contrast, ExpertDet achieves consistent improvements across all hierarchy levels, indicating that expert-knowledge guided supervision enhances model-level fine-grained discrimination while preserving semantic consistency at coarser levels. 
Therefore, ExpertDet provides more reliable detection results under different recognition granularities, better satisfying the multi-level recognition requirements in practical applications.

\begin{table*}[!t]
    \centering
    \caption{Comparison of ExpertDet with SOTA methods at different hierarchy levels on PSP.Ship and PSP.Plane.}
    \label{tab:coarse_results}
    \begin{tabular}{l|cccc|cccc}
        \toprule
        \multirow{2}{*}{Methods} & \multicolumn{4}{c|}{PSP.Ship} & \multicolumn{4}{c}{PSP.Plane} \\
        & $\mathrm{mAP}^0$ & $\mathrm{mAP}^1$ & $\mathrm{mAP}^2$ & $\mathrm{mAP}^3$ & $\mathrm{mAP}^0$ & $\mathrm{mAP}^1$ & $\mathrm{mAP}^2$ & $\mathrm{mAP}^3$ \\
        \midrule
        Oriented R-CNN~\cite{xie2021oriented} & 79.3 & 80.4 & 85.8 & 89.6 & 66.4 & 78.5 & 79.5 & 86.7 \\
        PCLDet~\cite{ouyang2023pcldet} & 78.4 \scriptsize{\myred{(-0.9)}} & 80.8 \scriptsize{\mygreen{(+0.4)}} & 86.7 \scriptsize{\mygreen{(+0.9)}} & 89.6 \scriptsize{\mygreen{(+0.0)}} & 66.3 \scriptsize{\myred{(-0.1)}}& 77.4 \scriptsize{\myred{(-1.1)}}& 79.4 \scriptsize{\myred{(-0.1)}}& 86.4 \scriptsize{\myred{(-0.3)}}\\
        EagleVision~\cite{jiang2025eaglevision} & 64.5 \scriptsize{\myred{(-14.8)}} & 66.0 \scriptsize{\myred{(-14.4)}} & 77.0 \scriptsize{\myred{(-8.8)}} & 86.4 \scriptsize{\myred{(-3.2)}} & 66.3 \scriptsize{\myred{(-0.1)}}& 78.2 \scriptsize{\myred{(-0.3)}}& 78.1 \scriptsize{\myred{(-1.4)}}& 86.2 \scriptsize{\myred{(-0.5)}}\\
        OM~\cite{zhu2024enhancing} & 77.9 \scriptsize{\myred{(-1.4)}} & 80.0 \scriptsize{\myred{(-0.4)}} & 85.9 \scriptsize{\mygreen{(+0.1)}} & 89.6 \scriptsize{\mygreen{(+0.0)}} & 66.3 \scriptsize{\myred{(-0.1)}}& 77.0 \scriptsize{\myred{(-1.5)}}& 79.1 \scriptsize{\myred{(-0.4)}}& 85.9 \scriptsize{\myred{(-0.8)}}\\
        \rowcolor{gray!20}
        ExpertDet & 82.5 \scriptsize{\mygreen{(+3.2)}} & 81.8 \scriptsize{\mygreen{(+1.4)}} & 88.2 \scriptsize{\mygreen{(+2.6)}} & 89.8 \scriptsize{\mygreen{(+0.1)}} & 67.4 \scriptsize{\mygreen{(+1.0)}}& 78.9 \scriptsize{\mygreen{(+0.4)}} & 79.9 \scriptsize{\mygreen{(+0.4)}} & 87.0 \scriptsize{\mygreen{(+0.3)}} \\
        \bottomrule
    \end{tabular}
\end{table*}